\documentclass[journal]{IEEEtran}

\usepackage{float}
\usepackage{amssymb}
\usepackage{amsmath}
\usepackage{graphicx}
\usepackage{booktabs}
\usepackage{multirow}
\usepackage[super]{nth}

\newcommand{\mat}[1]{\mathbf{#1}}
\newcommand{\superscript}[1]{\ensuremath{^\textrm{#1}}}
\DeclareMathOperator{\argmin}{\text{arg\,min}}

\begin{document}

\title{Spatiotemporal-Aware Augmented Reality: Redefining HCI in Image-Guided Therapy}

\author{Javad~Fotouhi, Arian~Mehrfard, Tianyu~Song, Alex~Johnson~M.D., Greg~Osgood~M.D., Mathias~Unberath, Mehran~Armand, and Nassir Navab
\thanks{J. Fotouhi, A. Mehrfard, T. Song, M. Unberath, M. Armand, and N. Navab are with the Laboratory for Computational Sensing and Robotics, Johns Hopkins University, Baltimore, MD, USA. \protect\\
E-mail: javad.fotouhi@jhu.edu, amehrfa1@jhu.edu, tsong11@jhu.edu, unberath@jhu.edu, marmand2@jhu.edu, nassir.navab@jhu.edu}
\thanks{A. Johnson, G. Osgood, and M. Armand are with the Department of Orthopaedic Surgery, Johns Hopkins Hospital, Baltimore,
MD, USA. \protect\\
E-mail: gosgood2@jhmi.edu, ajohn190@jhmi.edu, marmand2@jhu.edu}
\thanks{A. Mehrfard and N. Navab are with the Technical University of Munich, Munich, Germany. \protect\\
E-mail: arian.mehrfard@tum.de, nassir.navab@tum.de}
\thanks{J. Fotouhi, A. Mehrfard, and T. Song are regarded as joint first authors}% <-this % stops a space
\thanks{Manuscript Submitted to IEEE Transactions on Medical Imaging (T-MI)}}

\markboth{Journal of \LaTeX\ Class Files,~Vol.~XXX, No.~xxx, xxx~xxx}%
{Fotouhi \MakeLowercase{\textit{et al.}}: Spatiotemporal-Aware Augmented Reality}

\maketitle

\begin{abstract}
Suboptimal interaction with patient data and challenges in mastering 3D anatomy based on ill-posed 2D interventional images are essential concerns in image-guided therapies. Augmented reality (AR) has been introduced in the operating rooms in the last decade; however, in image-guided interventions, it has often only been considered as a visualization device improving traditional workflows. As a consequence, the technology is gaining minimum maturity that it requires to redefine new procedures, user interfaces, and interactions. The main contribution of this paper is to reveal how exemplary workflows are redefined by taking full advantage of head-mounted displays when entirely co-registered with the imaging system at all times. The proposed AR landscape is enabled by co-localizing the users and the imaging devices via the operating room environment and exploiting all involved frustums to move spatial information between different bodies. The awareness of the system from the geometric and physical characteristics of X-ray imaging allows the redefinition of different human-machine interfaces. We demonstrate that this AR paradigm is generic, and can benefit a wide variety of procedures. Our system achieved an error of $4.76\pm2.91$\,mm for placing K-wire in a fracture management procedure, and yielded errors of $1.57\pm1.16^\circ$ and $1.46\pm1.00^\circ$ in the abduction and anteversion angles, respectively, for total hip arthroplasty. We hope that our holistic approach towards improving the interface of surgery not only augments the surgeon's capabilities but also augments the surgical team's experience in carrying out an effective intervention with reduced complications and provide novel approaches of documenting procedures for training purposes.
\end{abstract}


\IEEEpeerreviewmaketitle

%%%%%%%%%%%%%%%%%%%%%%%%%%%%%%%%%%%%%%%%%%%%%%%%%%%%%%%%%%%%%
%%%%%%%%%%%%%%%%%%%%%%%%%%%%%%%%%%%%%%%%%%%%%%%%%%%%%%%%%%%%%

\section{Introduction}

Interventional image guidance is widely adopted across multiple disciplines of minimally-invasive and percutaneous therapies~\cite{siewerdsen2005volume, hott2004intraoperative, miller2010occupational, theocharopoulos2003occupational}. Despite its importance in providing anatomy-level updates, visualization of images and interaction with the intra-operative data are inefficient, thus requiring extensive experience to properly associate the content of the image with the patient anatomy. These challenges become evident in interventions that require the surgeon to navigate wires and catheters through critical structures under excessive radiation, such as in fracture or endovascular repairs.

Surgical navigation and robotic systems are developed to support surgery with localization and execution of well-defined tasks~\cite{kim2008use, kakarla2010placement, crawford2017surgical, yi2018robotic}. Though these systems increase the accuracy, their complex setup and explicit tracking nature may overburden the surgical workflow and consequently impede their acceptance in clinical routines~\cite{joskowicz2016computer}. Image-based navigation alleviates the requirements for external tracking, though depends strongly on pre-operative data which become outdated when the anatomy is altered during the surgery~\cite{grupp2019pose, goerres2017planning}. 

\begin{figure}[t]
  \centering
  \includegraphics[width=\columnwidth]{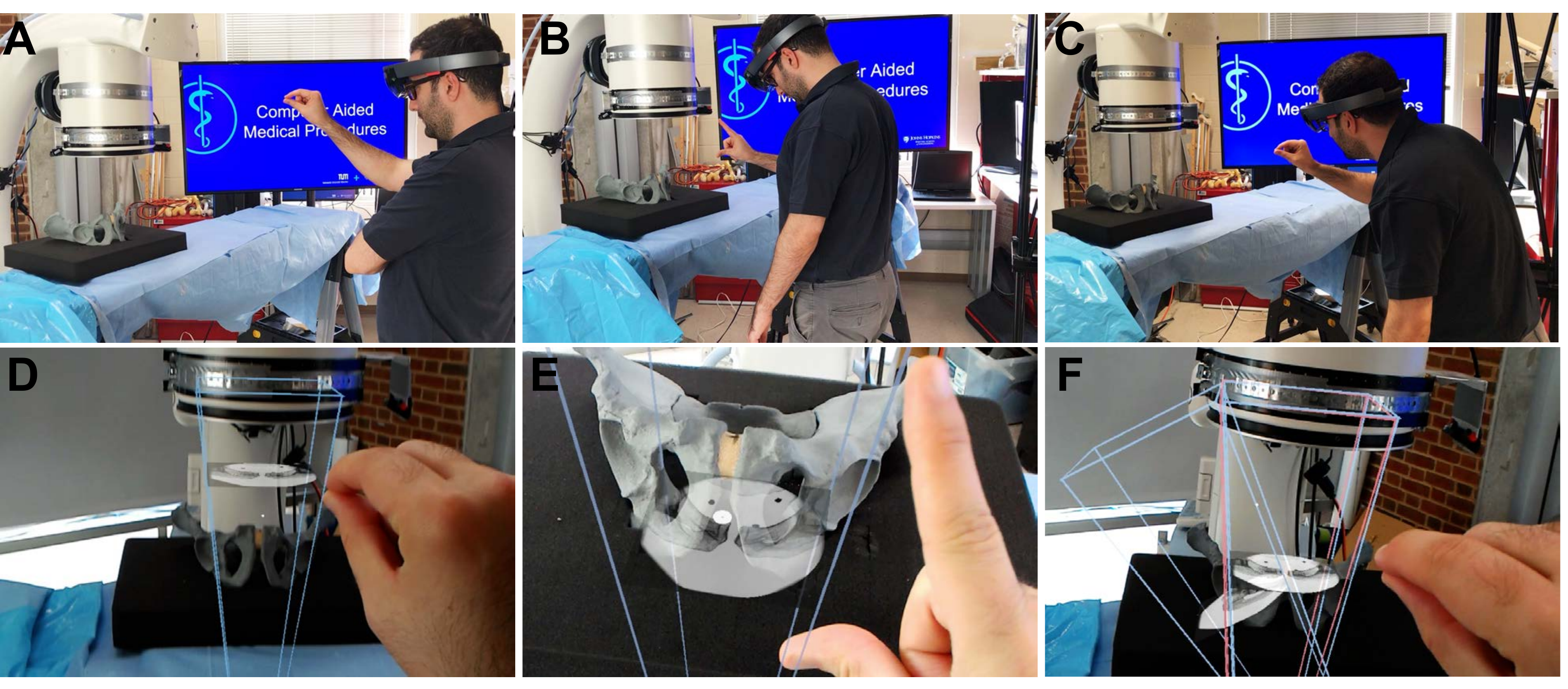}
  \caption{The augmented user interacts with the X-ray images within their viewing frustums (\textbf{A-C}). Corresponding AR views are shown in \textbf{D-F}.}
    \label{fig:frustums}
\end{figure}

As the surgical expectancy increases, the communication between the surgeon, crew, and the information becomes an important concern. Ineffective communication leads to increase of surgery time, radiation, and frustration to a point where in fluoroscopy-guided procedures, instead of the X-ray technician, the surgeons may reposition the scanners to ensure the task-defined views are optimal~\cite{synowitz2006surgeon, starr2001preliminary}.

To bridge the inefficiency gaps in surgical workflows, researchers have investigated the importance of human factor considerations in improving the usability of surgical data~\cite{aagaard2017interaction, laverdiere2019augmented}. Recent works focused on facilitating the unmet interaction needs by introducing touch-less mechanisms such as gaze, foot, or voice commands~\cite{hatscher2017gazetap, mewes2017touchless, ma2016device}. We believe the high stakes of surgery necessitates efficient interaction between all actors in the operating room \textit{i.e.} surgeon, anesthesiologist and staff to communicate and access information. This demands user-centric designs that can also accommodate fluid movement of information and surgical inference across the entire team.

Augmented reality (AR) solutions have gained popularity in computer-integrated surgeries, as they can provide intuitive visualizations of medical data directly at patients' site. Early works on surgical AR focused largely on multi-modal fusion of information and provided display-based overlays~\cite{sato1998image, navab1999merging, navab2009camera}. Subsequently, AR enabled the utilization of pre- and intra-interventional 3D data during therapies~\cite{fischer2016preclinical, fotouhi2016interventional, tucker2018towards, fotouhi2018plan}. 

The emergence of commercially available optical-see through head-mounted display (HMD) systems has led to development of AR solutions for various image-guided surgical disciplines, including percutaneous vertebroplasty, kyphoplasty, lumbar facet joint injection, orthopedic fracture management, bone cancer treatment, total hip arthroplasty (THA), interlocking nailing, cardiovascular surgeries, and surgical education~\cite{muller2019augmented,agten2018augmented,gibby2019head,van2018augmented,cho2018can,ogawa2018pilot,brun2018mixed,pelargos2017utilizing,laverdiere2019augmented}. 

AR has served as image viewer that directly displays the data at the operative site using virtual fluoroscopy monitors, hence eliminating the conventional off-axis visualization through static monitors~\cite{deib2018image, chimenti2015google}. Moreover, AR is used to provide navigational information during interventions~\cite{moreta2018augmented, meulstee2019toward, ma2018three}. These systems often rely on tracking of external markers, which require line-of-sight and invasive implantation into patients tissue that can hinder their usability. Andress et al. suggested a flexible marker-based surgical AR methodology which only required the marker to appear in the X-ray beam during the image acquisition, and was removed immediately after~\cite{andress2018fly}. Recent inside-out localization strategies in AR have greatly favored the fluid workflow over explicit navigation, and have proved effective in eliminating the need for external markers~\cite{hajek2018closing, fotouhi2019co, fotouhi2019interactive}.

\begin{figure}[t]
  \centering
  \includegraphics[width=\columnwidth]{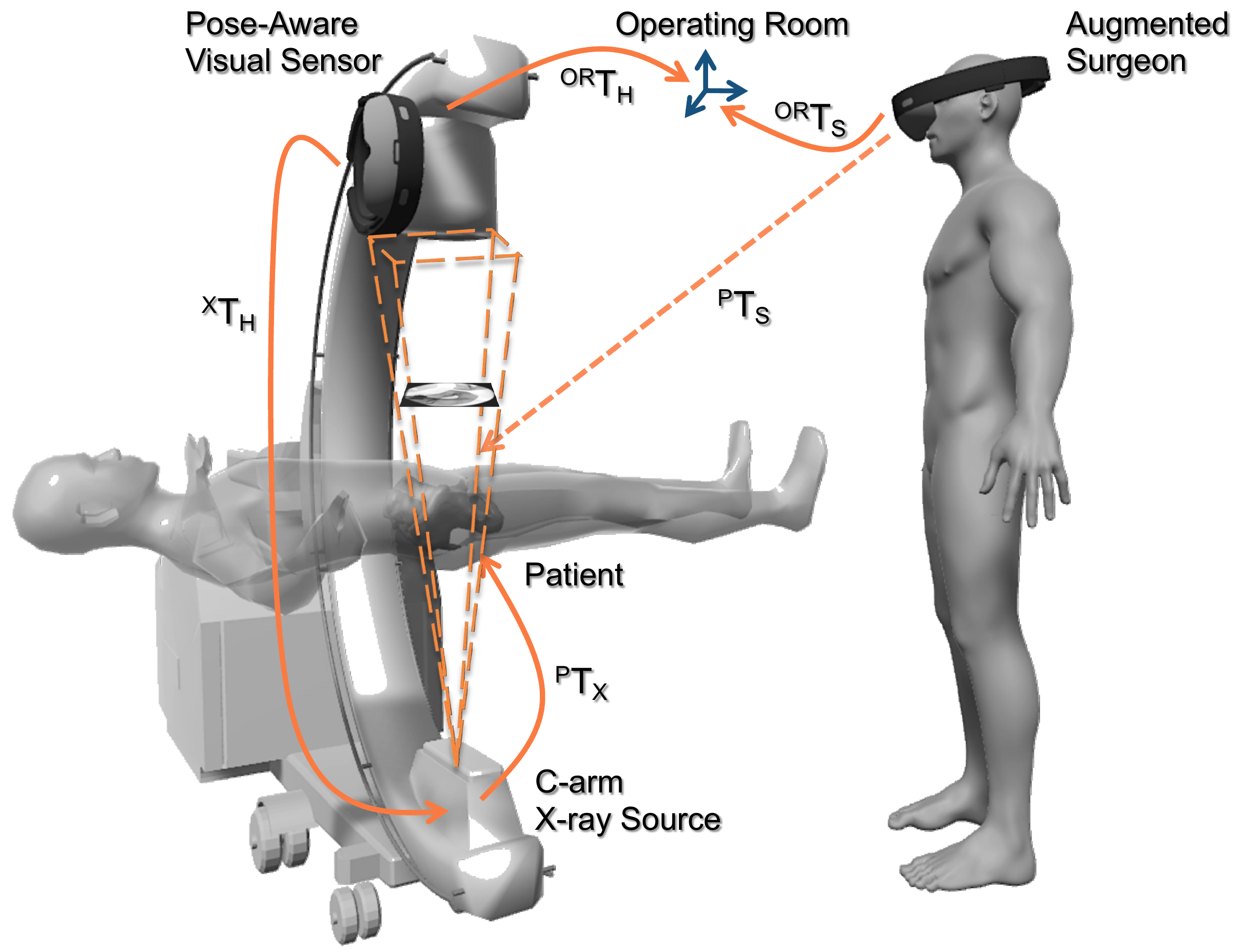}
  \caption{Transformation chain of the spatially-aware AR system}
    \label{fig:calibrationChain}
\end{figure}

This manuscript introduces the methodology and usability of a novel spatially-aware concept that enables immediate interaction with the medical data and promotes team approach where all stake-holders share a unified AR experience and communicate effectively. Our methodology exploits the viewing frustum of the imaging devices and human observers in the operating room (Fig.~\ref{fig:frustums}), and provides an engaging and immersive experience for the surgical team~\cite{fotouhi2019interactive}. We showcase this solution in two high volume orthopedic procedures, \textit{i.e.} K-wire placement in fracture care surgery, and acetabular cup placement in THA.

%%%%%%%%%%%%%%%%%%%%%%%%%%%%%%%%%%%%%%%%%%%%%%%%%%%%%%%%%%%%%
%%%%%%%%%%%%%%%%%%%%%%%%%%%%%%%%%%%%%%%%%%%%%%%%%%%%%%%%%%%%%

\section{Methodology}
Our main contributions are the collaborative AR concepts using spatiotemporal-aware flying frustums~\cite{fotouhi2019interactive} that enable intra-operative planning, define new workflows, support surgical crew, enhance the communication between surgeon and data, and enable intuitive documentation of the surgery for training purposes. We present the methodology for the realization of these concepts in the remainder of this section.

\subsection{Spatial-Awareness for AR}\label{subsec:spatialawarenss}
Visual data from cameras contain a wealth of information that can be used for simultaneous localization and mapping (SLAM). Visual SLAM is an important ingredient in our AR-based interaction recipe that enables co-localization of augmented users and the imaging device, which we will refer to as imaging observer, in a shared operating room environment. The relative pose between two frames $\alpha$ and $\beta$ using the environment map $\mat{M}$ can be estimated by minimizing the following reprojection function:
\begin{equation}
    \begin{aligned}
        ^{\alpha}\mathbf{T}_{\beta} &=~ \underset{\hat{^{\alpha}\mathbf{T}_{\beta}}}\argmin \;  D(\hat{^{\alpha}\mathbf{T}_{\beta}}, \mat{M}) \\
        &=~ \underset{\hat{^{\alpha}\mathbf{T}_{\beta}}}\argmin \sum_{f_i^{(\alpha)} \in I\alpha}
         | f_i^{(\alpha)} - P(\mat{M}(f_i^{(\alpha)})) |^2 \\
        &+~ \sum_{f_i^{(\beta)} \in I\beta} 
         | f_i^{(\beta)} - P\,\hat{^{\alpha}\mathbf{T}_{\beta}} (\mat{M}(f_i^{(\beta)})) |^2,
        \label{eq:slam}
    \end{aligned}
\end{equation}
where $f_i^{(\alpha)}$ and $f_i^{(\beta)}$ are corresponding features in images $I_\alpha$ and $I_\beta$, and $P$ is the projection operator.

In this setting, all users are localized with respect to a common spatial anchor in the operating room. The first member joining the shared experience will establish the anchor, \textit{i.e.} OR coordinate system, and every other member of the AR session will share their pose in a master-slave configuration with respect to this OR frame~\cite{hajek2018closing}. This relation is shown as $^{\text{OR}}\mathbf{T}_{\text{S}}$ in Fig.~\ref{fig:calibrationChain}.

\subsection{Imaging Observer}\label{subsec:imagingobserver}
C-arm scanners offer fluorscopic imaging capabilities for a wide range of less-invasive therapeutic areas. To seamlessly integrate this imaging device into our interactive AR paradigm, we augment the scanner with a rigidly attached visual tracker, that observers the structures in the OR environment, and communicates spatial information to all users. The materialization of this imaging observer system requires a co-calibration between the visual tracker on the scanner, and the X-ray source~\cite{fotouhi2019co}. The constant transformation that explains the calibration is denoted as $^\text{x}\mathbf{T}_\textbf{H}$ in Fig.~\ref{fig:calibrationChain}.

To estimate $^\text{x}\mathbf{T}_\textbf{H}$, we formulate an over-determined system of equations as follows:
\begin{equation}\label{eq:transformations}
    \begin{aligned}
        ^{\text{IR}}\mathbf{T}_{\text{OR}} &= ^{\text{IR}}\mathbf{T}_{\text{X}}(t_i) ^{\text{X}}\mathbf{T}_{\text{H}} {^{\text{OR}}\mathbf{T}^{-1}_{\text{H}}}  (t_i)\\
        &= ^{\text{IR}}\mathbf{T}_{\text{X}}(t_{i+1}) ^{\text{X}}\mathbf{T}_{\text{H}} {^{\text{OR}}\mathbf{T}^{-1}_{\text{H}}}  (t_{i+1}).
    \end{aligned}
\end{equation}
IR denotes the frame of an external tracker that is used to track the motion of the C-arm source as it undergoes different motion at times $t_i$ and $t_{i+1}$. It is important to note that the IR tracker is only used for this one-time and offline co-calibration, and it is not used intra-operatively. By re-arranging Eq.~\ref{eq:transformations}, we formulate the problem in the form of $\mat{A} \mat{X} = \mat{X} \mat{B}$ as presented in Eq.~\ref{eq:handeye}, such that $\mat{X} := ^{\text{X}}\mathbf{T}_{\text{H}}$, and $\mat{A}$ and $\mat{B}$ represent the relative motion of the X-ray source and the SLAM capable visual sensor on the gantry, respectively. 
\begin{equation}\label{eq:handeye}
        ^{\text{IR}}\mathbf{T}^{-1}_{\text{X}}(t_{i+1})
        ^{\text{IR}}\mathbf{T}_{\text{X}}(t_i) ^{\text{X}}\mathbf{T}_{\text{H}} 
        =  ^{\text{X}}\mathbf{T}_{\text{H}} {^{\text{OR}}\mathbf{T}^{-1}_{\text{H}}}  (t_{i-1})
        {^{\text{IR}}\mathbf{T}^{-1}_{\text{X}}}  (t_i).
\end{equation}
Rotation and translation components of the hand-eye problem are disentangled and computed separately as:
\begin{equation}\label{eq:handeye_decompose}
\begin{aligned}
& R_\mat{A} R_\mat{X} = R_\mat{X} R_\mat{B} \\
& R_\mat{A} \mat{t}_\mat{X} + \mat{t}_\mat{A} = R_\mat{X} \mat{t}_\mat{B} + \mat{t}_\mat{X}.
\end{aligned}
\end{equation}
We estimate the rotation parameters using unit quaternion representation as $q_\mat{A} \, q_\mat{X} = q_\mat{X} \,  q_\mat{B}$. Given that a unit quaternion $q_\mat{i}$ is formed by a vector $\mat{v}_\mat{i}$ and a scalar $s_\mat{i}$ such that $q_\mat{X} = \mat{v}_\mat{X} + s_\mat{X}$, we re-write the rotation component in Eq.~\ref{eq:handeye_decompose} using the quaternion product rule as:
\begin{equation}
    \begin{aligned}
    \vec{\mat{(.)}}: \, & s_\mat{A} \mat{v}_\mat{X} + s_\mat{X} \mat{v}_\mat{A} + \mat{v}_\mat{A} \times \mat{v}_\mat{X} = s_\mat{X} \mat{v}_\mat{B} + s_\mat{B} \mat{v}_\mat{X} + \mat{v}_\mat{X} \times \mat{v}_\mat{B}\\
    \tiny{(.):} \, & s_\mat{A} s_\mat{X} - \mat{v}_\mat{A} . \mat{v}_\mat{X} = s_\mat{X} s_\mat{B} - \mat{v}_\mat{X} . \mat{v}_\mat{B} .
    \end{aligned}
\end{equation}
Re-arranging the above formulation yields:
\begin{equation}
    \begin{bmatrix}
    s_\mat{A} - s_\mat{B} &  (\mat{v}_\mat{A} - \mat{v}_\mat{B})^\intercal \\
    (\mat{v}_\mat{A} - \mat{v}_\mat{B}) & (s_\mat{A} - s_\mat{B})I_3 + [\mat{v}_\mat{A} + \mat{v}_\mat{B}]_{\times}
    \end{bmatrix}
    \begin{bmatrix}
    s_\mat{X} \\ \mat{v}_\mat{X}
    \end{bmatrix} = 
    \begin{bmatrix}
    0 \\ \mat{0}_3
    \end{bmatrix},
\end{equation}
which is then solved in a constrained optimization fashion as:
\begin{equation}
\text{min} \, ||M\mat{q}||_2^2 \;\;\; s.t. \;\; ||\mat{q}||^2_2 = 1,
\end{equation}
where $\mat{q} = [ s_X, \vec{\mat{v}_x} ]^\intercal$. After the rotation parameters are computed, the translation vector is estimated in a least-squares setting: $(R_\mat{A} - I_3) \mat{t}_\mat{X} = R_\mat{X} \mat{t}_\mat{B} - \mat{t}_\mat{A}$.

\begin{figure}
  \centering
  \includegraphics[width=\columnwidth]{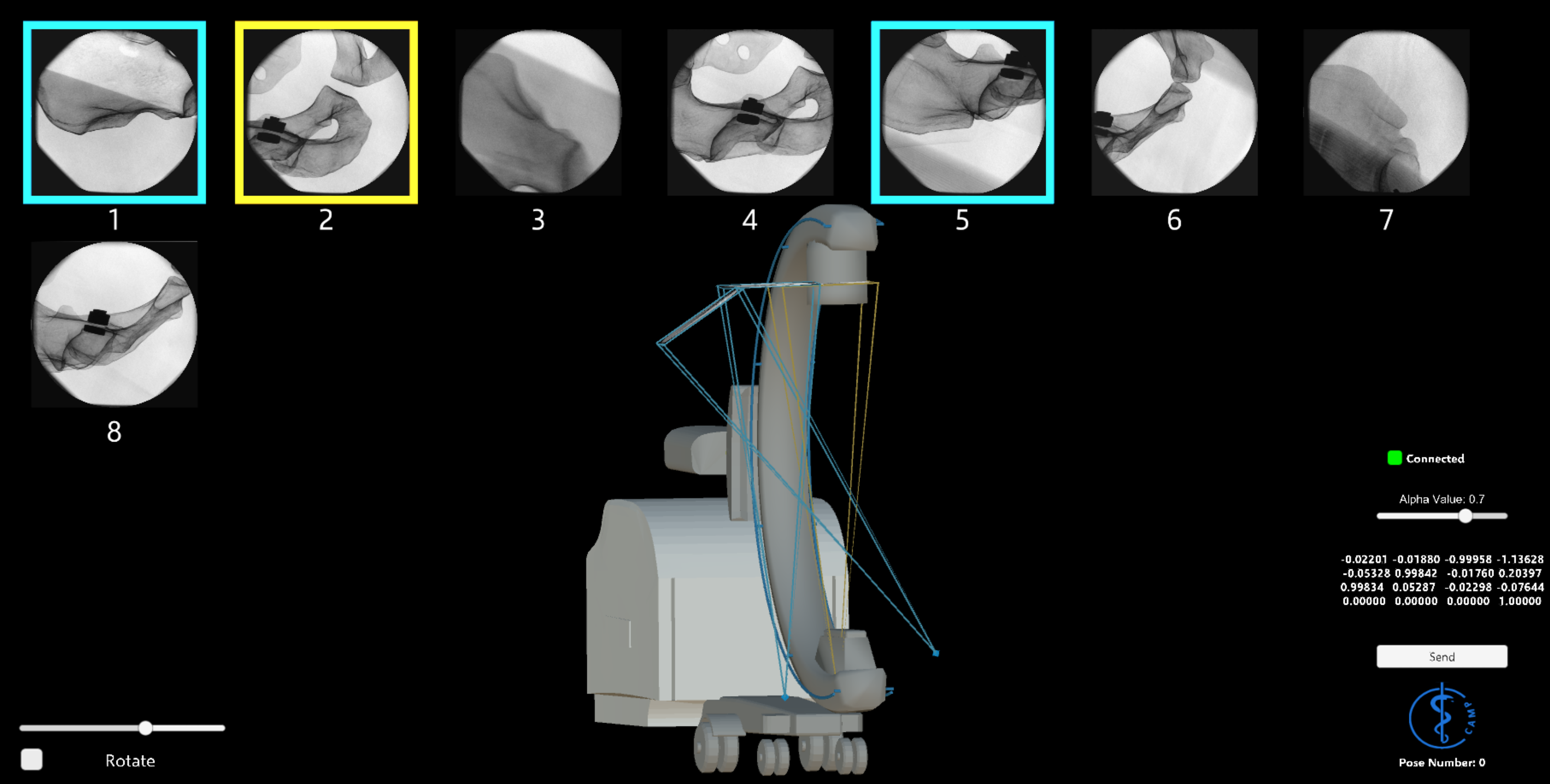}
  \caption{Multiple flying frustum are rendered at their corresponding 3D pose. The applications running on the computer and the HMD allows the users to replay various acquisition during or after the intervention.}
    \label{fig:App}
\end{figure}

\subsection{Geometry-Awareness for AR}\label{subsec:geometryawareness}
In this section, we describe the underlying geometry that allows us to combine the content of 2D X-ray images, directly with the 3D spatial information we computed in Sec.~\ref{subsec:spatialawarenss} and~\ref{subsec:imagingobserver}. To this end, we explicitly model the viewable region of the X-ray camera, known as the flying frustum~\cite{fotouhi2019interactive}, and allow interaction with images within their geometries. It is important to note that, the flying frustum refers to the full pyramid of vision (Fig.~\ref{fig:calibrationChain}), and is different than the truncated pyramids used in the computer graphics community. Despite the similarities in formulation, the conventional frustum model in graphics only applies to reflective images, and cannot accommodate the transmission model used in fluoroscopy. Therefore, we extend the perspective pinhole camera model that is commonly used in the computer vision community~\cite{hartley2003multiple}. 

\begin{figure}
  \centering
  \includegraphics[width=\columnwidth]{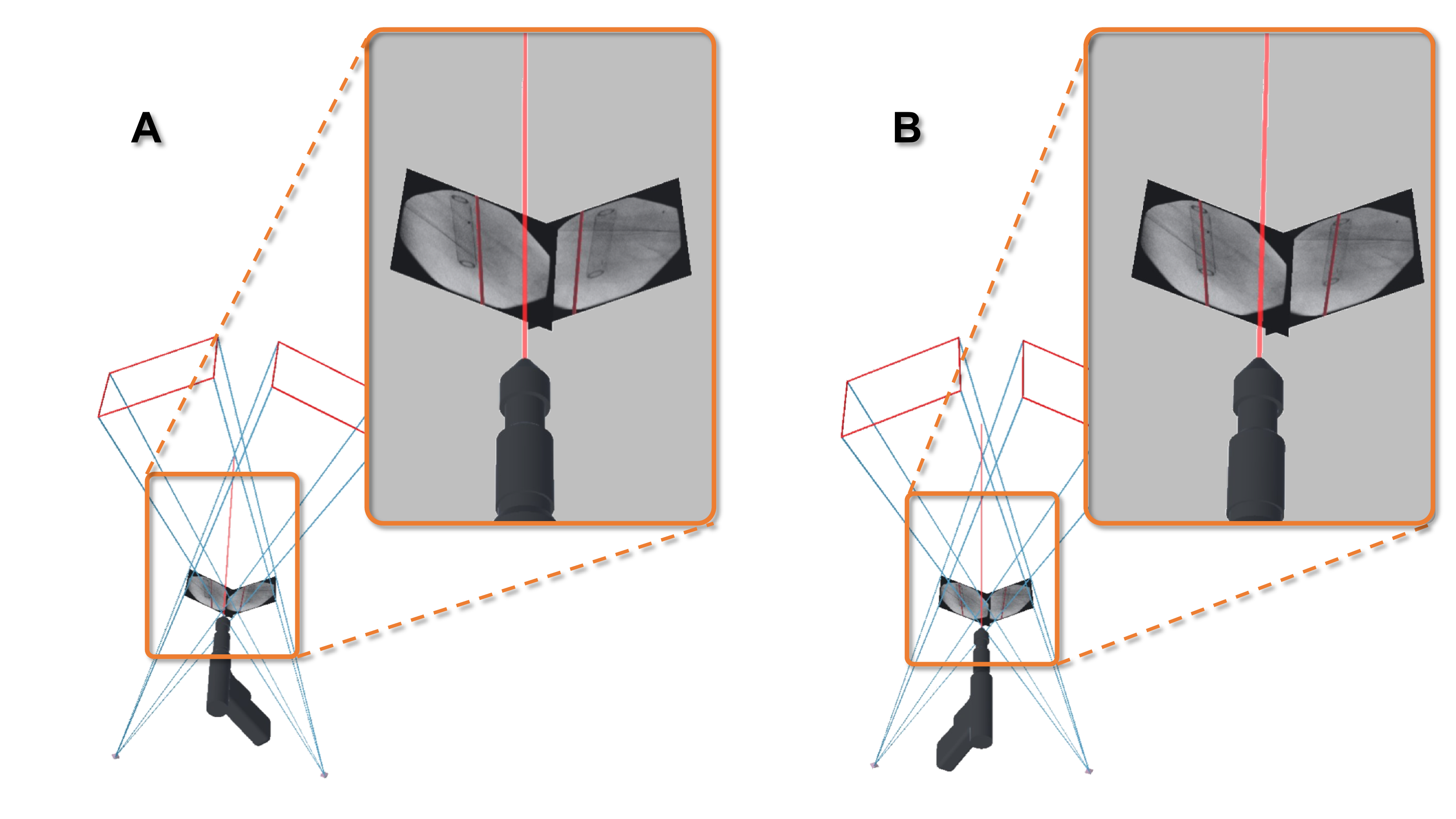}
  \caption{The augmented projections allow us to exploit the geometry in AR and plan surgical tools in relation to patient anatomy. The misaligned virtual drill in \textbf{A} is repositioned until it appears inside the desired structure in all the frustums (\textbf{B}).}
    \label{fig:CubeAlignment}
\end{figure}

In our paradigm, users can move the images within their frustums on a virtual plane known as the \textit{near} plane, between X-ray source and detector (referred to as the \textit{far} plane), while they remain a valid image of the same anatomy. This interaction enables the users to intersect the images with corresponding anatomies, and intuitively observe \textit{2D-image-to-3D-anatomy} associations. Additionally, the imaging technologists which operate the scanner, can align the scanner with a desired frustum that is decided by the surgeon.

A flying frustum is defined using the following model:
\begin{equation}\label{eq:mirror}
    P_{f} = 
    \begin{bmatrix} 
        \frac{n}{f} & 0 & 0 \\ 
        0 & \frac{n}{f} & 0 \\
        0 & 0 & 1  
    \end{bmatrix} \,
    K \,
    P \,
    \begin{bmatrix}
        ^{\text{OR}} R _{\text{X}} & ^{\text{OR}} \mat{t} _{\text{X}} \\
        \mat{0}^\top & 1
        \end{bmatrix},
\end{equation}
where $n$ refers to the distance to the \textit{near} plane, $f$ is the focal length, $K$ is the matrix of intrinsic parameters, and $0 \leq n \leq f$. The parameter $n$ is controlled by the user, such that when $n=f$, the X-ray image is directly displayed at the detector scale. It is worth mentioning that, with conventional frustum models, the \textit{near} plane can only take values smaller than the \textit{far} plane, which is not the case in our representation. 

\begin{figure}
  \centering
  \includegraphics[width=\columnwidth]{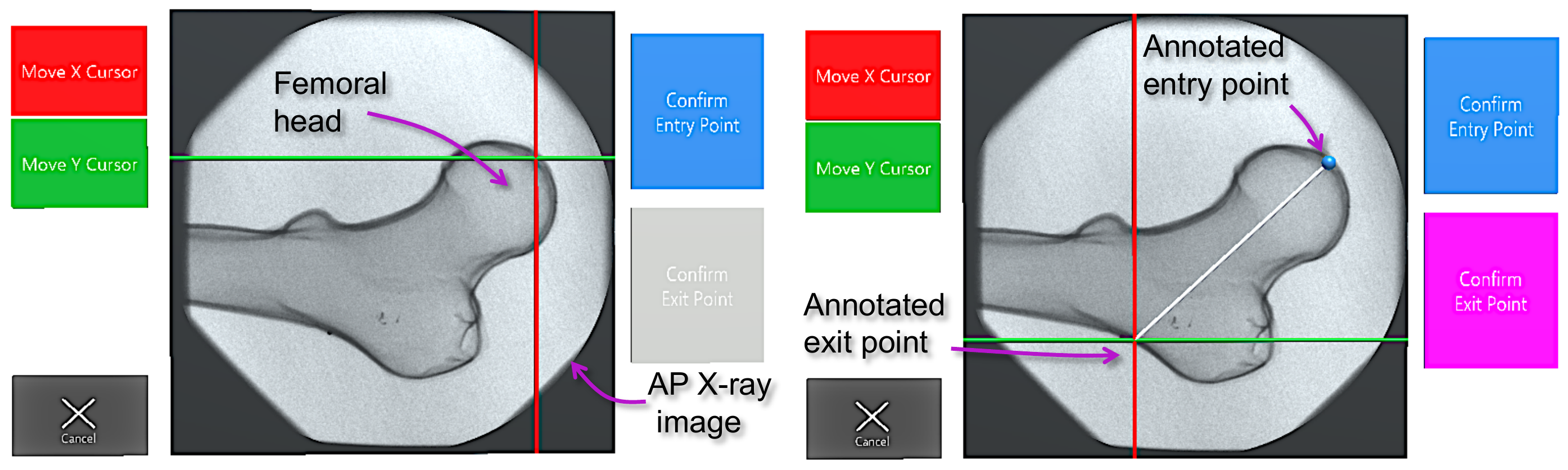}
  \caption{AR interface to plan trajectories on the X-ray acquisitions}
    \label{fig:entryExitPoint}
\end{figure}

\begin{figure}
  \centering
  \includegraphics[width=\columnwidth]{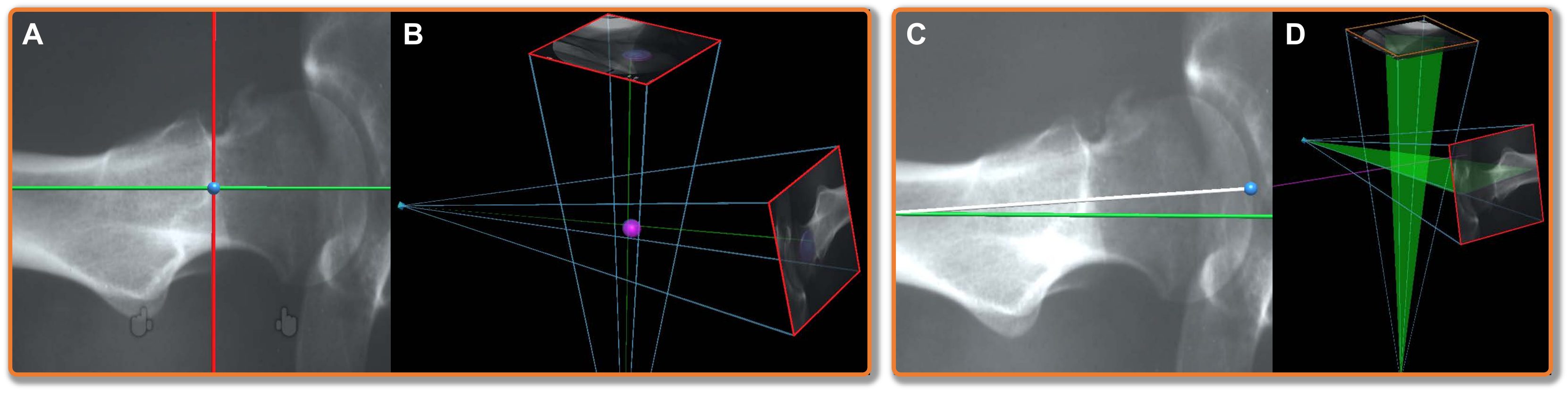}
  \caption{Each point in a frustum image corresponds to a ray passing through the landmark in 3D, and connecting the source and detector of the C-arm. Intersection of two rays recovers the 3D point and renders it directly on the patient (\textbf{A-B}). Similarly, annotation of lines in each frustum, corresponds to a plane in 3D. The intersection of these planes restores the 3D planning trajectory, and renders it in AR such that it travels through the corresponding anatomical structure (\textbf{C-D}).}
    \label{fig:intersection}
\end{figure}

For each 2D point $x_i \in \mathcal{I}$, where $\mathcal{I}$ is the domain of all acquired images, the corresponding point $x_f$ in the frustum domain $\mathcal{F}$ is scaled by a factor $s$ as $\mat{x}_f = s \, \mat{x}_i = (\frac{n}{f}) \mat{x}_i\, , \; \text{such that} \; 0 \leq n \leq f$.
Finally, the 3D pose of the interactive image in the frustum is defined as:
\begin{equation}
    \begin{aligned}
        ^{\text{OR}}\mathbf{T}_{\text{I}} &=
        \begin{bmatrix}
            ^{\text{OR}} R _{\text{X}} & ^{\text{OR}} \mat{t} _{\text{X}} \\
            \mat{0}^\top & 1
        \end{bmatrix}
        \begin{bmatrix}
            I_3 & \begin{matrix} 0 \\ 0 \\ n \end{matrix} \\
            \mat{0}^\top & 1
        \end{bmatrix} \\
        &= 
        \begin{bmatrix}
            ^{\text{OR}} R _{\text{X}} & \begin{matrix} r_{13}\, n \\ r_{23}\, n \\ r_{33}\, n \end{matrix} + ^{\text{OR}} \mat{t} _{\text{X}} \\
            \mat{0}^\top & 1
        \end{bmatrix},
    \end{aligned}
\end{equation}
where $R = \left\{r_{i,j}\right\}_{i,j : 1,\,2,\,3}$. Fig.~\ref{fig:App} demonstrates multiple flying frustums, each rendered given their respective 3D pose.

\begin{figure}
  \centering
  \includegraphics[width=\columnwidth]{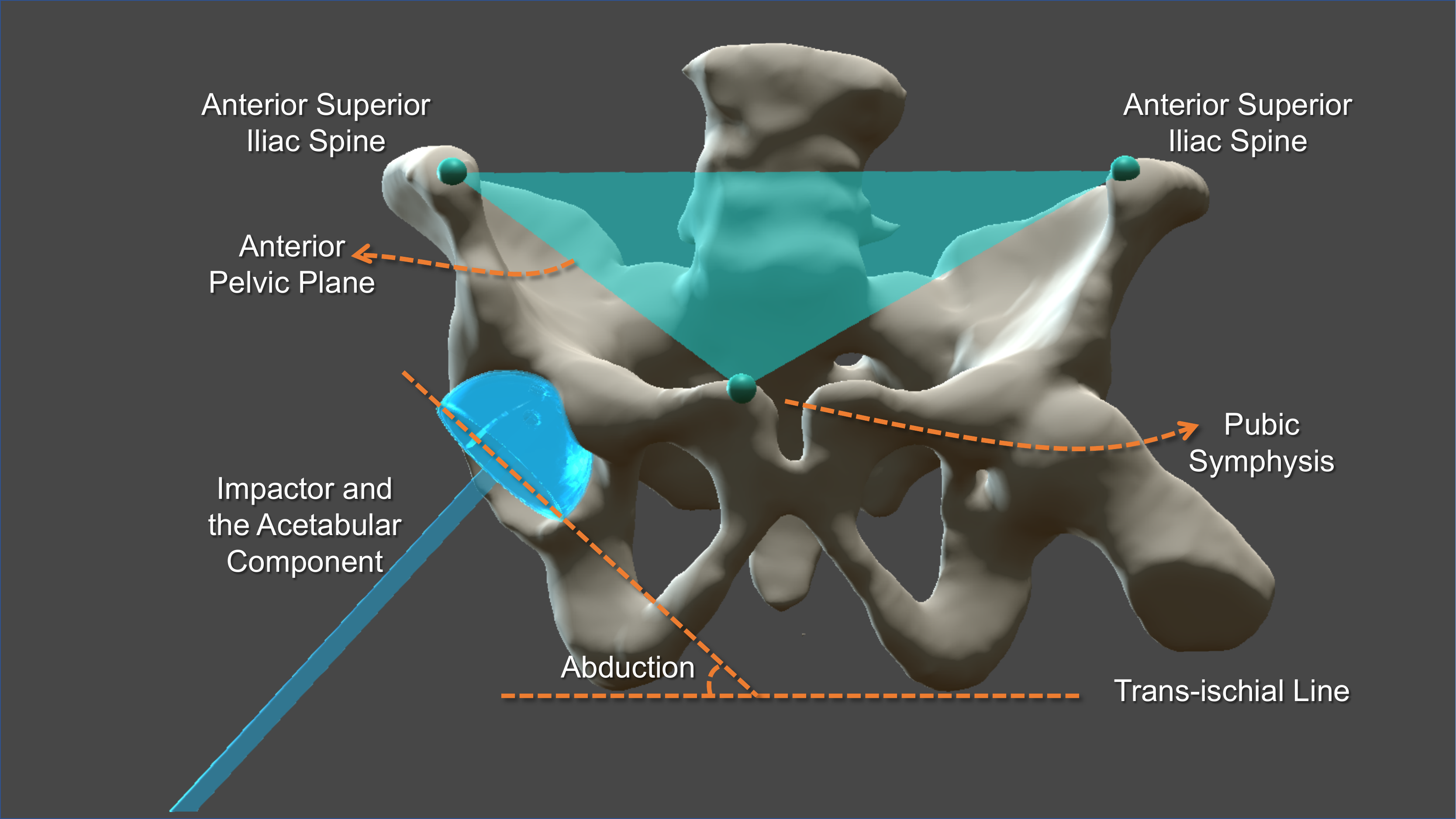}
  \caption{In THA, abduction and anteversion angles of the acetabular implant are defined with respect to the anterior pelvic plane.}
    \label{fig:APPplane}
\end{figure}

\begin{figure*}
  \centering
  \includegraphics[width=\textwidth]{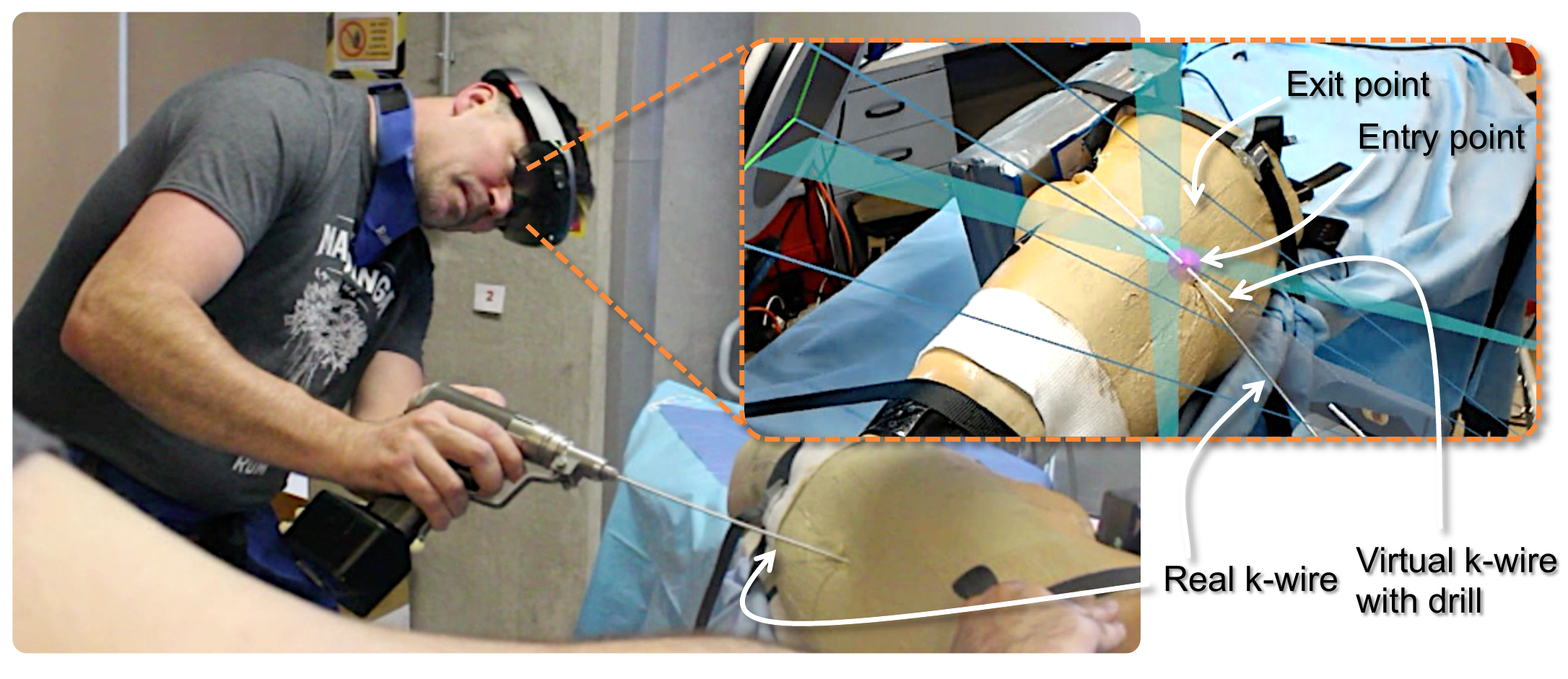}
  \caption{K-wire placement in a fracture reduction procedure}
    \label{fig:osgoodView}
\end{figure*}

\subsection{Planning Using Flying Frustums}\label{subsec:virtualprojection}
Flying frustums discussed in Sec.~\ref{subsec:spatialawarenss}-\ref{subsec:geometryawareness} embed sufficient 3D and 2D information that enable interventional planning for the placement of surgical tools. In this section we introduce two distinct approaches for intra-operative planning.

In the first method, virtual tools are manipulated in 3D by the user, and simultaneously projected onto the X-ray images of all valid frustums. A point $X_t \in \mathcal{T}$, where $\mathcal{T}$ is the domain of all 3D points on a virtual tool, is projected onto the $i^{\text{th}}$ frustum as $x_{t_i} = P_{f_i} X_t$.

In an exemplary case shown in Fig.~\ref{fig:CubeAlignment}, the virtual drill is rotated and translated until it passes through a desired structure (e.g. through a bone canal) in all frustums. An alignment consensus in all frustums is the equivalent of the alignment of the virtual 3D tool with the imaged anatomy in 3D. 

The second planning approach requires 2D interaction on the frustum X-ray images. In this setting, for each selected landmark on a frustum image (Fig.~\ref{fig:entryExitPoint}), a 3D ray connecting the C-arm source and the target landmark will be rendered into the AR scene. As illustrated in Fig.~\ref{fig:intersection}, the intersection of two rays from a corresponding landmark in two images reconstruct the 3D landmark. Each ray is defined via two elements: \textit{i)} the position of the C-arm X-ray source $\mat{c}_i$, and \textit{ii)} the unit direction vector $\mat{u}_i$ from the source to the annotated landmark in the frustum. We estimate the closest point $\mat{x}_l^{*}$ to the $N=2$ rays corresponding to each landmark $l$ via a least-squares minimization strategy as follows:
\begin{equation}
    \begin{aligned}
        & \mat{x}_l^{*} = \underset{\mat{x}\in\mathbb{R}^3}\argmin{ \sum_{i=1}^N \Vert (I_3 -\mat{u}_i \mat{u}_i^{\top})\mat{x} - \mat{t}_i\Vert^2}, \\
        & \text{where} \, \, \, \mat{t}_i= (I_3 - \mat{u}_i \mat{u}_i^{\top})\mat{c}_i\,.
    \end{aligned}
\end{equation}

Similarly, two points on a frustum $i$ defining the entry and the exit points of a drilling trajectory, associate to two rays $\mat{u}_{1i}$ and $\mat{u}_{2i}$ in 3D. These two rays span a plane in 3D as shown in Fig.~\ref{fig:intersection}. The intersection of the planes corresponding to the same entry and exit points on frustums $i$ and $j$ form a 3D line $\mat{d}_{12} = (\mat{u}_{1i} \times \mat{u}_{2i}) \times (\mat{u}_{1j} \times \mat{u}_{2j})$ that passes through the desired entry and exit points on the patient anatomy.

Our first approach requires a more complex interaction with the augmented surgical implant using the $6$ \textit{degrees-of-freedom}, however generalizes to arbitrary structures beyond linear annotations, such as the curved plates used for internal fixations.

\subsection{Surgical Workflow Integration}\label{subsec:surgicalusecase}
Intra-operative planning and execution with the flying frustums support can be used in various fluoroscopy-guided procedures. In THA, the critical points defining the anterior pelvic plane (APP) can be each identified on X-ray images. Given APP, a virtual acetabular implant and a rigidly attached impactor are rendered in AR with their desired orientation that is calculated with respect to APP. Likewise, the translational component of the cup implant is identified by defining the center of the patient acetabulum on corresponding fluoroscopic images. These relations are shown in Fig.~\ref{fig:APPplane}.

Another exemplary image-guided procedure is the placement of screws and K-wires during fracture management. As shown in Fig.~\ref{fig:osgoodView}, AR provides support for placement of K-wires using the trajectory planning on the corresponding frustums.

%%%%%%%%%%%%%%%%%%%%%%%%%%%%%%%%%%%%%%%%%%%%%%%%%%%%%%%%%%%%%
%%%%%%%%%%%%%%%%%%%%%%%%%%%%%%%%%%%%%%%%%%%%%%%%%%%%%%%%%%%%%

\section{Experimental results}
\subsection{System Setup}

Our system comprises an ARCADIS Orbic 3D C-arm (Siemens Healthineers, Forchheim, Germany) as an intra-operative X-ray device that automatically computes the cumulative area dose for each session. The immersive AR solution was built using the Unity cross-platform game engine (Unity Technologies, San Francisco, CA, US) and was deployed to an optical-see-through HMD, the Microsoft HoloLens (Microsoft, Redmond, WA). To jointly co-localize the augmented surgeon and the C-arm scanner, a second HoloLens device with inside-out SLAM capabilities was attached near the X-ray detector. The two HMDs shared their spatial anchor, a rich feature reference region in the common environment, over a wireless local network, allowing them to remain synchronized and establish spatial awareness. This connection was enabled through a TCP-based sharing service running on an Alienware (Dell, Round Rock, TX, US) laptop server with an Intel i7-7700HQ CPU, NVIDIA GTX 1070 graphics card, 16 GB RAM, and Windows 10 operating system. The X-ray images from C-arm were transmitted to the server computer over a direct Ethernet connection, and then uploaded to the HMD.

\subsection{System Calibration}
\begin{figure}
  \centering
  \includegraphics[width=\columnwidth]{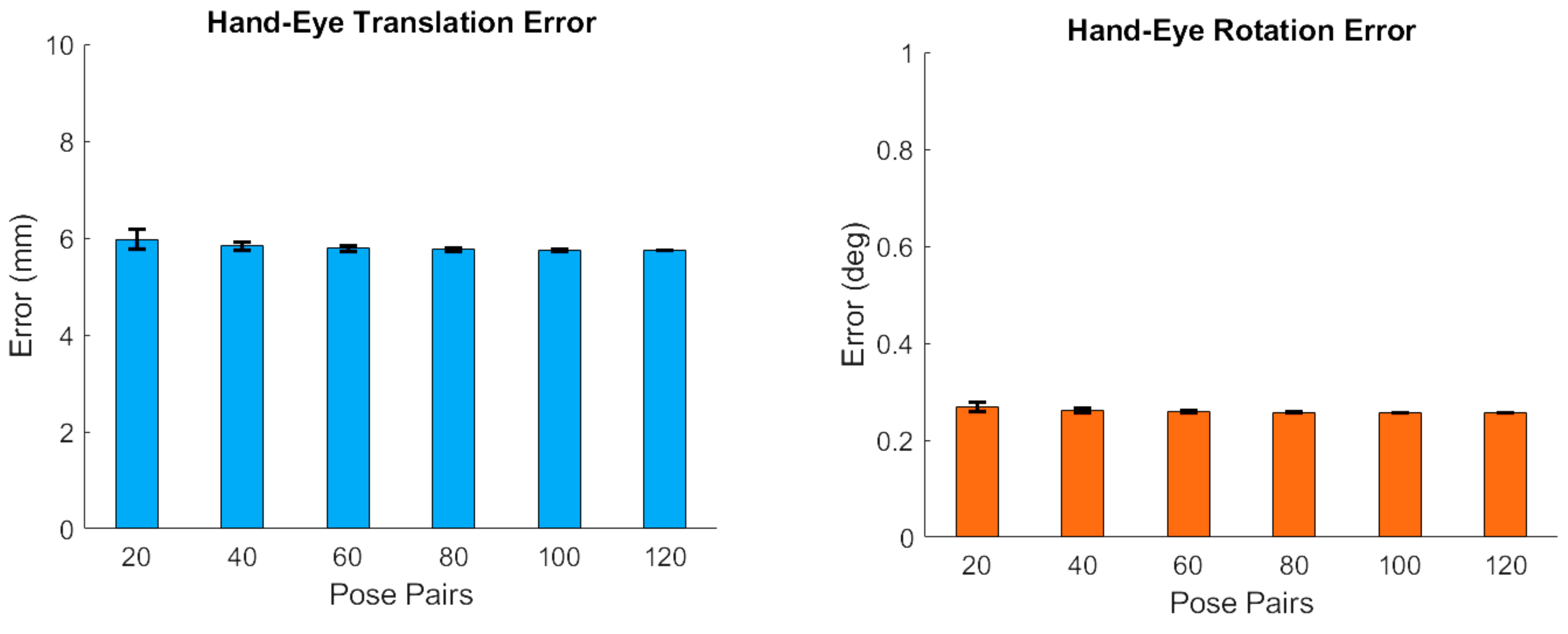}
  \caption{Mean and standard deviation of the translational and rotational errors for the hand eye calibration step are shown in the left and right plots, respectively. For each $N$ number of pose pairs shown on the horizontal axis (except the last column which considers all the available data), the experiments were repeated $100$ times by each time sampling $N$ poses from the total $120$ available poses and computing the hand-eye calibration.}
    \label{fig:handeyeError}
\end{figure}

To solve the hand-eye calibration problem in Eq.~\ref{eq:handeye}, $120$ different pairs of corresponding poses were recorded from the visual tracker on the C-arm as well as an external infrared tracking system that tracked the C-arm source. Fig.~\ref{fig:handeyeError} presents the error for this offline calibration step given different sampling for the pose pairs.

The localization quality of the SLAM-based visual tracking system on the scanner, \textit{i.e.} the HMD on the scanner, was compared to a ground-truth provided by an external tracker. We measured rotational errors of $(0.71^\circ, 0.11^\circ, 0.74^\circ)$ with a norm of $0.75^\circ$ and translational errors of $(4.0, 5.0, 4.8)$~mm with a norm of $8.0$~mm along the $(x, y, z)$ axes, respectively~\cite{fotouhi2019interactive}.

\subsection{Experiments}
Eight orthopedic surgeons and residents from Johns Hopkins Hospital participated in pre-clinical user studies and performed two surgically relevant tasks while utilizing interactive flying frustums in an immersive AR environment.

\begin{figure}
  \centering
  \includegraphics[width=\columnwidth]{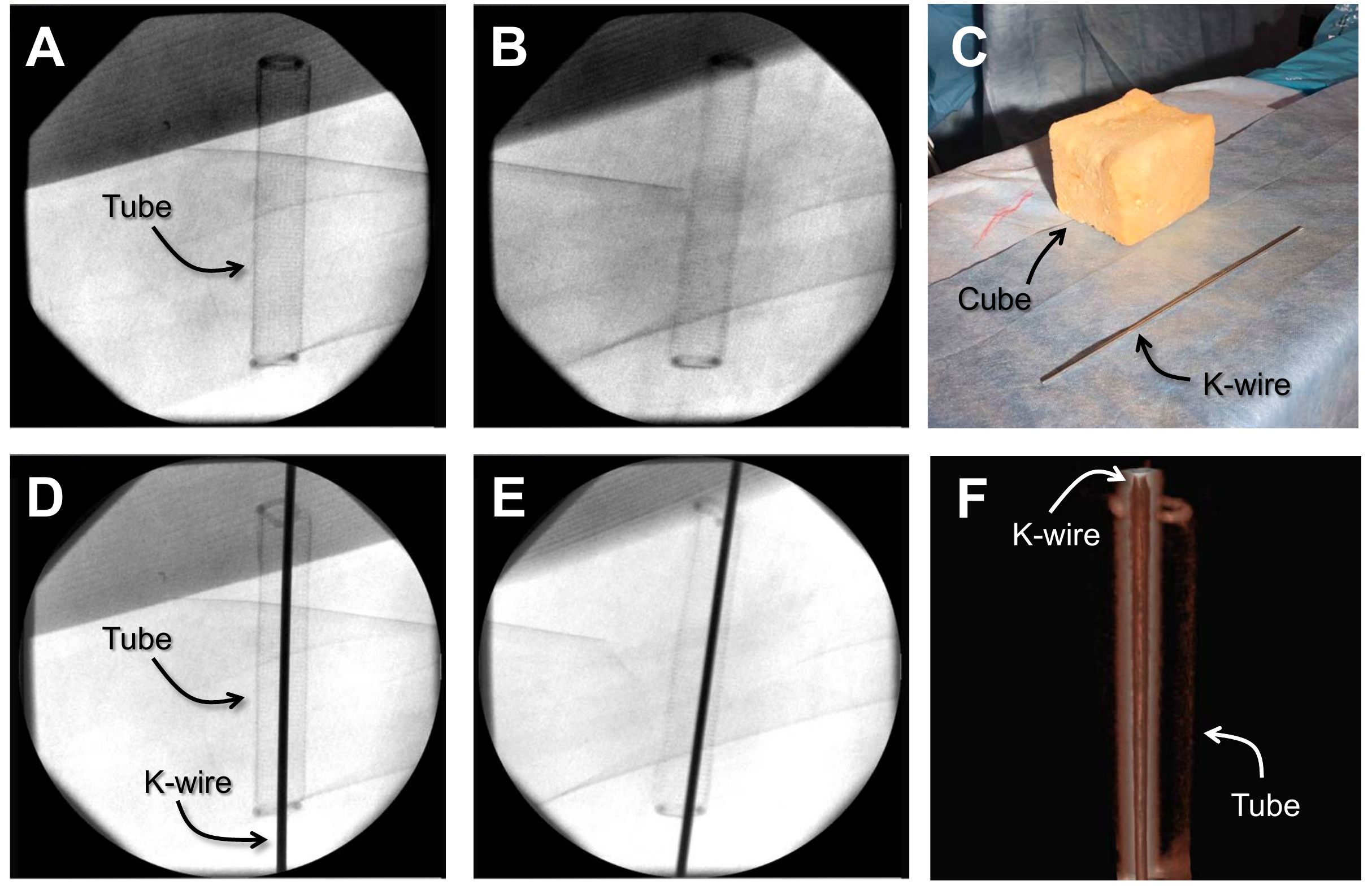}
  \caption{\textbf{A-B} are the X-ray images of the cubic phantom shown in \textbf{C}. In \textbf{D-E}, the X-ray images of the same phantom are shown after a K-wire was successfully inserted inside the tube. \textbf{F} is the CBCT scan of the phantom which was acquired for verification. Due to metal artifacts, the tube does not exhibit strong contrast.}
    \label{fig:cube}
\end{figure}

In the first procedure, we focused on the correct placement of a K-wire to repair complex fractures. To emulate the K-wire placement through the superior pubic ramus (acetabulum arc), we used radiopaque cubic phantoms, as seen in Fig. \ref{fig:cube}-\textbf{C}. For direct comparison, we used the same setup that was used by Fischer et al.~\cite{fischer2016preclinical}. Each cube consisted of a stiff, lightweight, and non radiopaque methylene bisphenyl diisocyanate (MDI) foam. Since the superior pubic ramus is a tubular bone with a diameter of approximately $10$~mm, we used a thin aluminium mesh filled with MDI that was placed inside each cube and served as the bone phantom. The two ends of the tubular structures were complemented with a rubber radiopaque ring. Each subject was asked to place a K-wire with a diameter of $2.8$~mm through the tubular phantom using a surgical drill (Stryker Corporation, Kalamazoo, MI, US).

\begin{figure}
  \centering
  \includegraphics[width=\columnwidth]{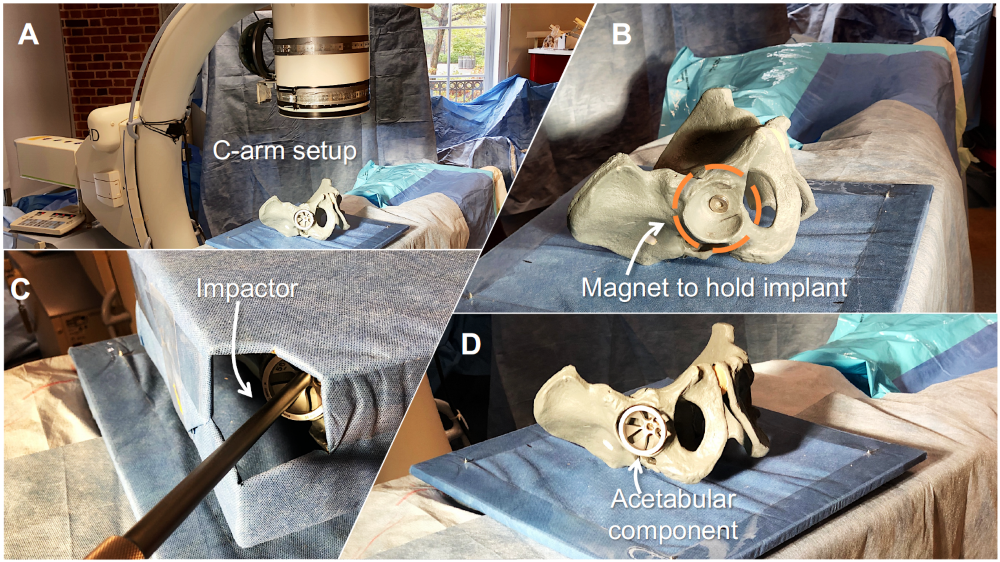}
  \caption{In \textbf{A} the setup of the C-arm, pelvic phantom, and the acetabular cup are shown. \textbf{B} is a close-up view of the phantom with an empty acetabular socket and a magnet for holding the implant in position. Image \textbf{C} shows the impactor while it is placed by a surgeon during the experiment, and \textbf{D} shows the successfully placed cup in the acetabulum.}
    \label{fig:tha}
\end{figure}

For the second procedure, we constructed a total hip arthroplasty mock setup by using a radiopaque pelvis phantom with a magnetic acetabulum to fixate the acetabular cup (Fig.~\ref{fig:tha}). For direct comparison, we adopted the same experimental setup that was suggested by Alexander et al.~\cite{alexander2020augmented}. The cup was attached to a straight cylindrical acetabular trialing impactor (Smith \& Nephew, London, UK) allowing the operator to guide the cup. Since the ideal orientation of the implant is unknown, we use abduction and anteversion angles that lie in a safe zone defined by landmarks on the pelvis as described in~\cite{lewinnek1978dislocations}.

Initially, each surgeon received a brief introduction to the Microsoft Hololens, preparing them to properly mount and use the HMD. To further instruct them on our AR application, pre-recorded training X-ray images were loaded onto their HMD, allowing them to become familiar with the interface, planning procedure, and the interaction mechanism using hand gestures. 

After the required planning images were acquired by the proctors, each surgeon planned their respective procedure in AR and performed the drilling task into the cube or placed the acetabular component into the pelvis. During the procedure, they were explicitly allowed to order as many X-ray shots from any perspective that they considered necessary. 

We recorded the planing time, the time it took them to execute the procedure, number of fluoroscopic acquisitions, and the cumulative radiation dose as it was measured by the scanner. Finally, for the verification and accuracy measurement, we acquired a 3D cone-beam CT (CBCT) scan of the phantoms with their respective implants.

\begin{table*}
\centering
\caption{Outcome from K-wire insertion for participants $\mathbf{Pi}$}
\label{tab:Kwire}
\begin{tabular}{lcccccccc}
 & $\mathbf{P1}$ & $\mathbf{P2}$ & $\mathbf{P3}$ & $\mathbf{P4}$ & $\mathbf{P5}$ & $\mathbf{P6}$ & $\mathbf{P7}$ & $\mathbf{P8}$ \\ \hline
\textbf{Planning Time (sec)} & $125$ & $40$ & $39$ & $58$ & $79$ & $22$ & $67$ & $44$ \\
\textbf{Execution Time (sec)} & $83$ & $74$ & $58$ & $46$ & $66$ & $67$ & $63$ & $79$ \\
\textbf{\# X-ray images} & $2$ & $2$ & $2$ & $2$ & $2$ & $2$ & $2$ & 2 \\
\textbf{Dose (cGY(cm\textsuperscript{2}))} & $0.28$ & $0.21$ & $0.19$ & $0.27$ & $0.26$ & $0.28$ & $0.27$ & $0.28$ \\
\textbf{Error (mm)} & $8.23$ & $5.71$ & $9.02$ & $3.26$ & $6.94$ & $1.13$ & $1.59$ & $2.23$ \\ \hline
\end{tabular}
\end{table*}

\begin{table*}
    \centering
    \caption{Outcome from the placement of the acetabular implant for participants $\mathbf{Pi}$}
    \label{tab:THA}
    \begin{tabular}{lcccccccc}
         & $\mathbf{P1}$ & $\mathbf{P2}$ & $\mathbf{P3}$ & $\mathbf{P4}$ & $\mathbf{P5}$ & $\mathbf{P6}$ & $\mathbf{P7}$ & $\mathbf{P8}$ \\ \hline
        \textbf{Planning Time (sec)} & $162$ & $70$ & $117$ & $88$ & $64$ & $37$ & $71$ & $110$ \\
        \textbf{Execution Time (sec)} & $87$ & $39$ & $13$ & $19$ & $17$ & $35$ & $26$ & $24$ \\
        \textbf{\# X-ray images} & $8$ & $8$ & $8$ & $8$ & $8$ & $8$ & $8$ & $8$ \\
        \textbf{Dose (cGY(cm\textsuperscript{2}))} & $1.27$ & $1.3$ & $1.23$ & $1.26$ & $1.18$ & $1.25$ & $1.18$ & $1.29$ \\
        \textbf{Abduction error (\textsuperscript{$\circ$})} & $2.1$ & $1$ & $1.1$ & $1.3$ & $2.9$ & $0.1$ & $0.4$ & $3.7$ \\
        \textbf{Anteversion error (\textsuperscript{$\circ$})} & $1.1$ & $0.6$ & $2.7$ & $1.4$ & $2.1$ & $0.4$ & $0.3$ & $3.1$ \\ \hline
    \end{tabular}
\end{table*}

\subsection{Results}
Tables~\ref{tab:Kwire} and~\ref{tab:THA} comprise the performance of every participant in the experiments. Table~\ref{tab:Kwire} contains the measurements for the K-wire insertion, and Table~\ref{tab:THA} presents the procedural outcome for the acetabular cup placement. We separate the interventional time measurements into \textit{i)} \textit{planning time}, the time it took each surgeon to plan their procedure in AR, and \textit{ii)} \textit{execution time}, determining the duration of the insertion/placement of the instruments. 
Furthermore, we recorded the number of X-ray acquisitions and the respective dose for each user. 
Finally, to assess the overall performances, we computed the average distance of the K-wire from the center of the tube at the entry and the exit surface of the tubular structure, and the abduction and anteversion angles of the acetabular implant, based on standard guidelines.

\begin{figure}
  \centering
  \includegraphics[width=0.95\columnwidth]{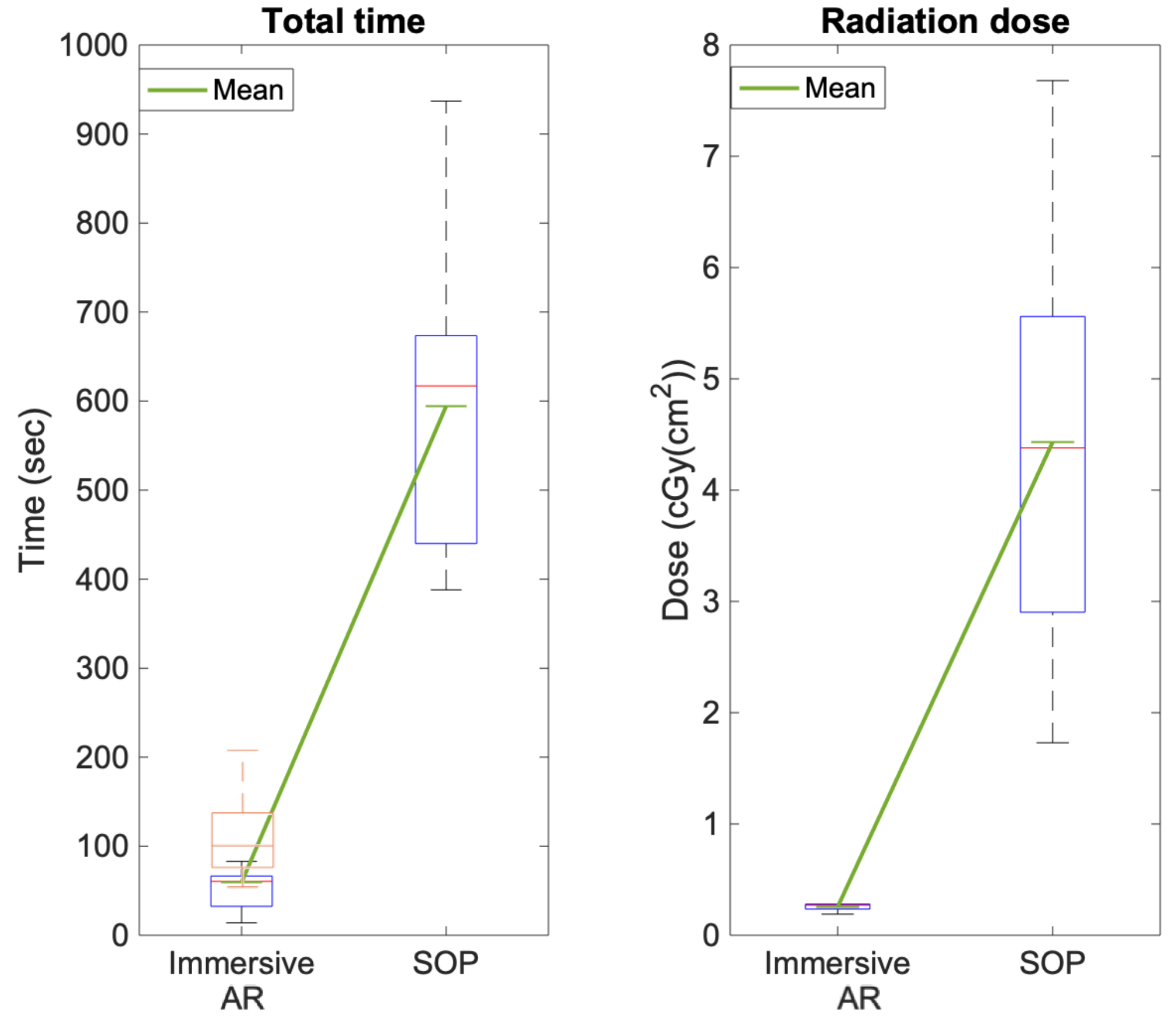}
  \caption{The plots present the execution time and total radiation dose during K-wire insertion using the AR supported approach and SOP. On the leftmost plot, the blue boxplot is the execution time with AR, whereas the orange boxplot is the total time including the planning phase. The green lines show the mean values for each of the groups.}
    \label{fig:cubeBoxplot}
\end{figure}

Table~\ref{tab:KwireComp} compares the K-wire insertion results of our immersive AR system with a previous non-immersive AR system~\cite{fischer2016preclinical} as well as the standard operating procedure (SOP) using conventional fluoroscopic guidance. Combining the planning and execution times, the AR procedure took on average $111.25$\,sec versus the $594.3$\,sec during SOP. Fig.~\ref{fig:cubeBoxplot} depicts this comparison.
On average, the surgeons used $2$ fluoroscopic shots with a combined dose of $0.255$\,(cGY(cm\textsuperscript{2})) per user and committed an insertion error of $4.76$\,mm.
The associated standard deviation (SD) values are presented in Table~\ref{tab:KwireCompSDEV}. 

\begin{table}
    \caption{Comparison of AR, non-immersive AR (NI-AR), and SOP for K-wire insertion. We denote the time for the AR procedure as ($A + B$), with $A$ being the planning time and $B$ the execution time.}
    \label{tab:KwireComp}
    \centering
    \begin{tabular}{lccccc}
\textbf{MEAN} & \textbf{Time} & \textbf{\# X-ray} & \textbf{Dose} & \textbf{Error} \\
 & \textbf{(sec)} & \textbf{images} & \textbf{(cGY(cm\textsuperscript{2}))} & \textbf{(mm)} \\ \hline
\textbf{AR} & $59.25+52$ & $2$ & $0.255$ & $4.76$ \\
\textbf{NI-AR}\tiny{~\cite{fischer2016preclinical}} & $243.7$ & $2.14$ & $1.6$ & $5.13$ \\
\textbf{SOP} & $594.3$ & $40.86$ & $4.43$ & $4.61$ \\ \hline
    \end{tabular}
\end{table}

\begin{table}
    \caption{Performance during acetabular implant placement using AR, NI-AR, and SOP. We denote the time for the AR procedure as ($A + B$), with $A$ being the planning and $B$ the execution time.}
    \label{tab:THAComp}
    \centering
    \begin{tabular}{lccccc}
\textbf{MEAN} & \textbf{Time} & \textbf{\# X-ray} & \textbf{Dose} & \textbf{Abd.} & \textbf{Ant.} \\
 & \textbf{(sec)} & \textbf{images} & \textbf{(cGY(cm\textsuperscript{2}))} & \textbf{(\textsuperscript{$\circ$})} & \textbf{(\textsuperscript{$\circ$})} \\ \hline
\textbf{AR} & $89.88+32.5$ & $8$ & $1.25$ & $1.57$ & $1.46$ \\
\textbf{NI-AR}\tiny{~\cite{alexander2020augmented}} & $180$ & $1$ & $1.83$ & $1.78$ & $1.43$ \\
\textbf{SOP} & $235$ & $13.75$ & $1.96$ & $4.76$ & $4.77$ \\ \hline
\end{tabular}
\end{table}

\begin{figure}
  \centering
  \includegraphics[width=0.95\columnwidth]{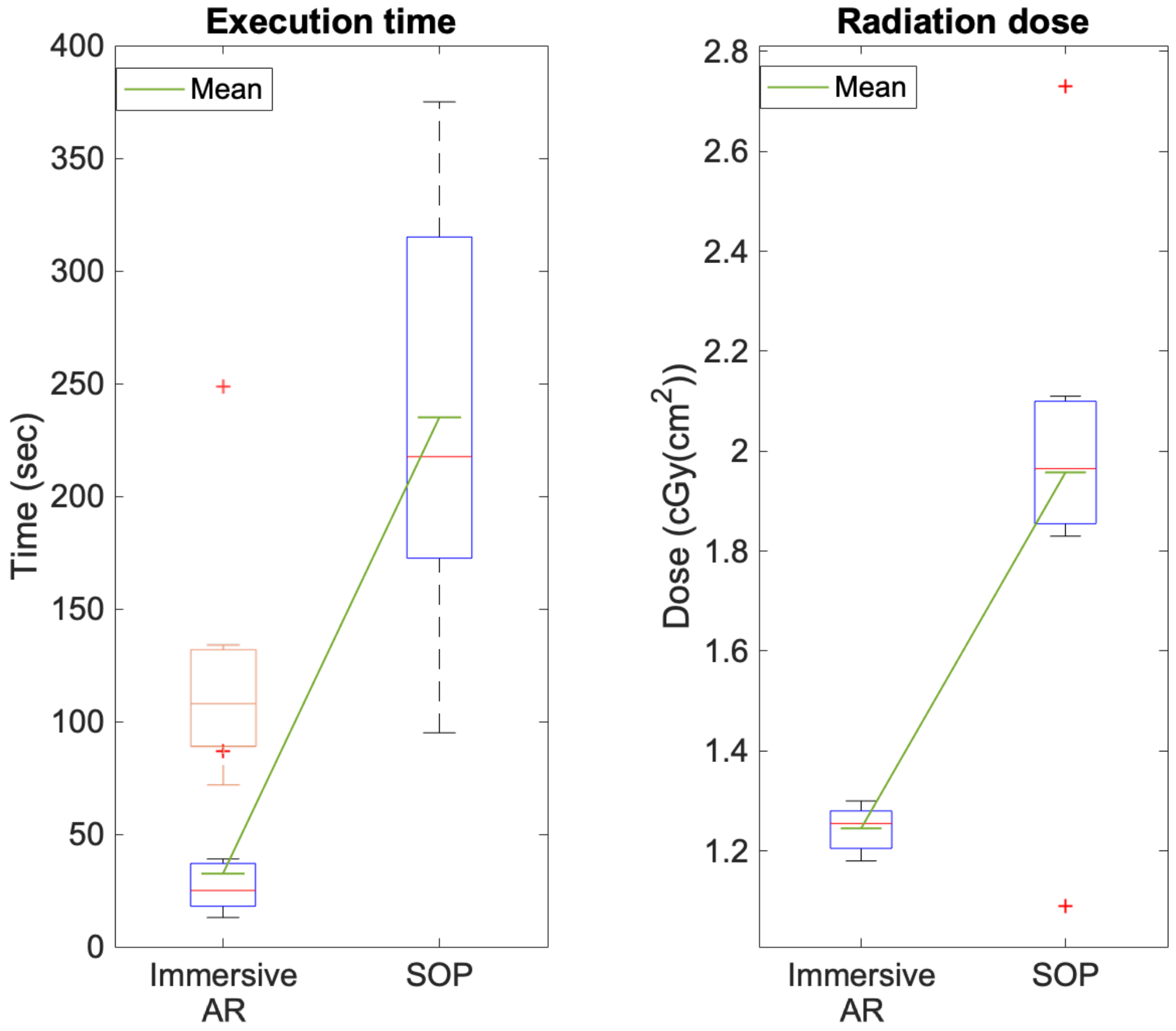}
  \caption{Comparison of time and total radiation dose during cup placement with AR and SOP approaches. The orange boxplot represents the total time including the planning time. The red ($+$) denote outliers, where in the leftmost plot the top sign belongs to the orange boxplot, and the bottom ($+$) to the blue plot.}
    \label{fig:thaBoxplot}
\end{figure}

\begin{figure}
  \centering
  \includegraphics[width=0.9\columnwidth]{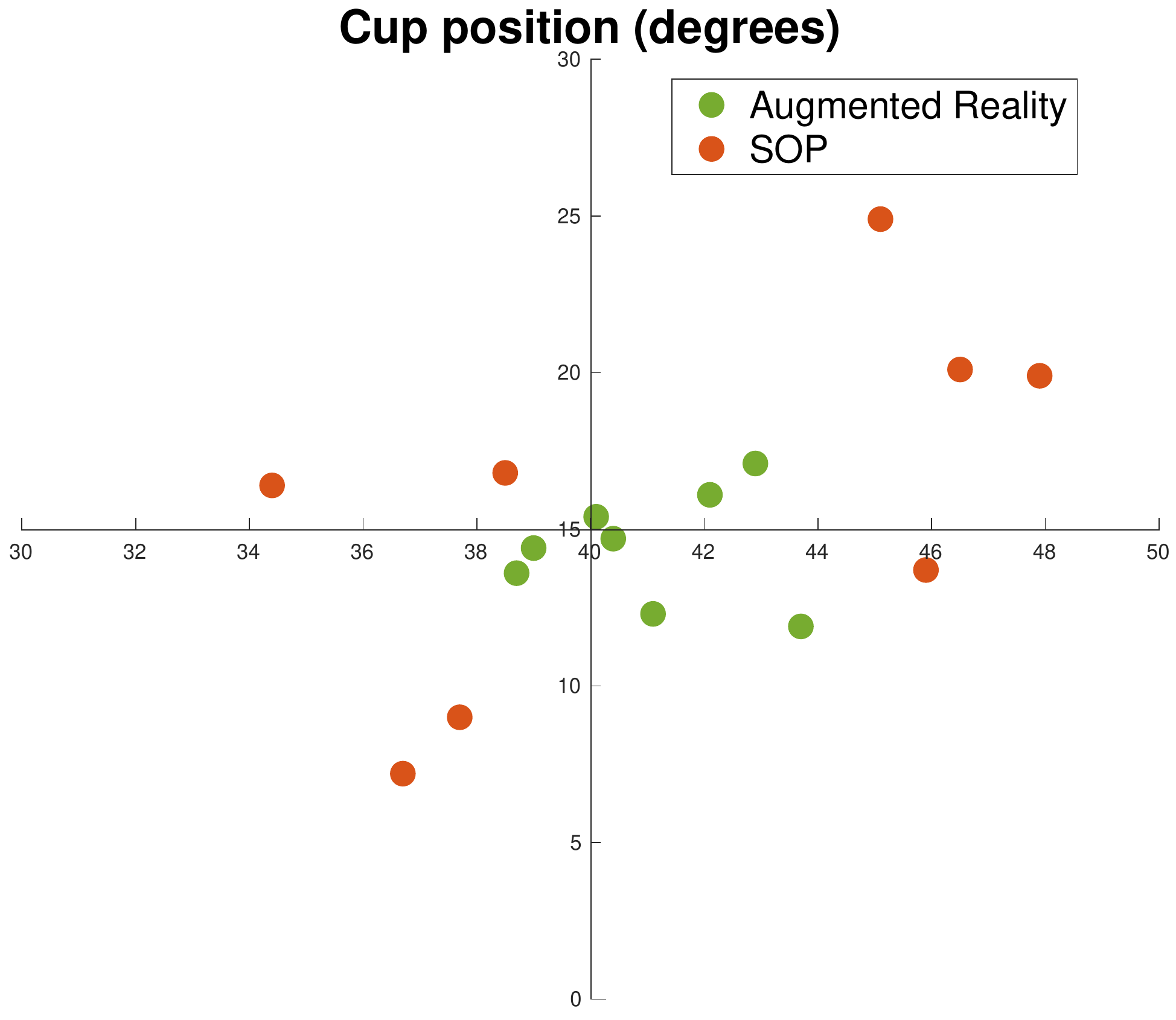}
  \caption{Anteversion and abduction angles after acetabular cup placement using AR support and SOP}
    \label{fig:thaScatter}
\end{figure}

In Table~\ref{tab:THAComp} we present the outcome for the acetabular cup placement procedure with the immersive AR system, comparing it to a previous non-immersive AR application~\cite{alexander2020augmented} and SOP. Using SOP, it took surgeons on average $235$\,sec to place the cup and under AR a combined time of $122.38$\,sec was achieved. For the AR setup, we acquired $8$ X-ray images with an average dose of $1.25$\,(cGY(cm\textsuperscript{2})) per surgeon, whereas $14$ fluoroscopic images with a dose of $1.96$\,(cGY(cm\textsuperscript{2})) were acquired during SOP. With the AR system, the average errors were $1.57^\circ$ and $1.46^\circ$ for the abduction and anteversion angles, respectively. Under SOP the respective angles were $4.76^\circ$ and $4.77^\circ$. Figs.~\ref{fig:thaBoxplot} and~\ref{fig:thaScatter} present the outcome with respect to time, radiation dose, and individual rotational measures for the acetabular cup placement experiments using AR and SOP.
The corresponding SD are listed in Table~\ref{tab:THACompSDEV}. The immersive AR results show an SD of respectively $89.88$ sec and $32.5$\,sec for planning and execution time, $0$ for the number of X-ray images, $1.25$ for the dose, $1.24$ for the abduction error and $1.07$ in the anteversion error. 

\begin{table}
    \caption{The respective standard deviation values, relating to Table~\ref{tab:KwireComp}.}
    \label{tab:KwireCompSDEV}
    \centering
    \begin{tabular}{lccccc}
\textbf{SD} & \textbf{Time} & \textbf{\# X-ray} & \textbf{Dose} & \textbf{Error} \\
 & \textbf{(sec)} & \textbf{images} & \textbf{(cGY(cm\textsuperscript{2}))} & \textbf{(mm)} \\ \hline
\textbf{AR} & $32.02+24.23$ & $0$ & $0.04$ & $3.11$ \\
\textbf{NI-AR}\cite{fischer2016preclinical} & $84.00$ & $0.69$ & $0.17$ & $2.72$ \\
\textbf{SOP} & $188.0$ & $19.38$ & $2.00$ & $3.62$ \\ \hline
\end{tabular}
\end{table}

\begin{table}
\centering
\caption{The respective standard deviation values, relating to Table~\ref{tab:THAComp}.}
\label{tab:THACompSDEV}
\begin{tabular}{lccccc}
\textbf{SD} & \textbf{Time} & \textbf{\# X-ray} & \textbf{Dose} & \textbf{Abd.} & \textbf{Ant.} \\
 & \textbf{(sec)} & \textbf{images} & \textbf{(cGY(cm\textsuperscript{2}))} & \textbf{(\textsuperscript{$\circ$})} & \textbf{(\textsuperscript{$\circ$})} \\ \hline
\textbf{AR} & $89.88+32.5$ & $0$ & $1.25$ & $1.24$ & $1.07$ \\
\textbf{NI-AR}\cite{alexander2020augmented} & $15$ & $0$ & $0.06$ & $1.37$ & $0.66$ \\
\textbf{SOP} & $96$ & $3.73$ & $0.45$ & $2.2$ & $3.15$ \\ \hline
\end{tabular}
\end{table}

%%%%%%%%%%%%%%%%%%%%%%%%%%%%%%%%%%%%%%%%%%%%%%%%%%%%%%%%%%%%%
%%%%%%%%%%%%%%%%%%%%%%%%%%%%%%%%%%%%%%%%%%%%%%%%%%%%%%%%%%%%%

\section{Discussion}

We evaluated our spatially-aware AR system in two clinically relevant procedures, \textit{i)} the placement of K-wires through tubular structures for fracture repair tasks,  and \textit{ii)} placement of acetabular components into the hip socket for total hip arthroplasty. We selected these two high volume procedures among the many other applications which can be enabled by our interactive AR system, as they each represent a class of common orientational alignment and localization tasks that are prevalent across different fields of image-guided surgery.

For the K-wire insertion procedure, the immersive AR system performed substantially faster than the conventional SOP, yielding less than a fifth of the time (Fig.~\ref{fig:cubeBoxplot}). Table~\ref{tab:KwireComp} demonstrates a detailed comparison of our system, not only with the SOP as an established baseline, but also with a previously presented non-immersive mixed reality method based on RGBD sensing and intra-operative CBCT imaging~\cite{fischer2016preclinical}.

With the AR system every surgeon used exactly $2$ X-ray images, which were the $2$ images required for procedure planning. Despite explicitly allowing them to take as many radiographs as they desire, no one of the surgeons requested additional X-ray images. As mentioned above, during SOP, surgeons inserted the K-wires with an average of $40.86$ fluoroscopic images and with an average dose of $4.43$\,cGY(cm\superscript{2}), compared to $0.255$ cGY(cm\superscript{2}), which were used during the AR procedures. The RGBD-CBCT system in~\cite{fischer2016preclinical, fotouhi2016interventional} yielded on average $2.14$ X-rays, although it required a pre-procedural CBCT scan of the phantom, inducing the higher radiation dose of $1.6$ cGY(cm\superscript{2}).

Finally, evaluating the outcome of the procedure with regard to the drilling error, AR ($4.76$\,mm) outperforms RGBD-CBCT ($5.13$\,mm), both being marginally worse than SOP ($4.61$\,mm). Considering that we only instructed the surgeons to drill through the tube and not precisely through the center of the tube, we regard these difference as negligible. It is important to note that, our AR system performed similar to the conventional X-ray method in terms of accuracy, while reducing time by a factor of $5$, number of fluoroscopic acquisitions by a factor of $20$, and the radiation dose by a factor of $17$. 

The same trend is observed with the measurements for the placement of the acetabular cup, demonstrating the effectiveness of our AR system, we compare it against SOP and a NI-AR system as presented in \cite{alexander2020augmented}. As shown in Fig.~\ref{fig:thaBoxplot}, the execution time is considerably lower using AR; even when combining planning and execution time it took the surgeons $122.38$\,sec, which is nearly half of the $235$\,sec that they needed under SOP and less than the $180$\,sec with NI-AR. Furthermore, the number of fluoroscopic images were reduced; every surgeon used exactly $8$ images, which are again merely the images required for planning. This resulted in an average dose of $1.25$\,cGY(cm\superscript{2}), which is lower than with SOP, where the surgeons used an average of $13.75$ radiographs with an average dose of $1.96$\,cGY(cm\superscript{2}), and with NI-AR where one pre-procedure CBCT lead to a dose of $1.83$\,cGY(cm\superscript{2}). The objective of this procedure was to achieve abduction and anteversion angles of $40^\circ$ and $15^\circ$, respectively, which lie in the clinical safe-zone~\cite{lewinnek1978dislocations}. The respective errors are shown in Table~\ref{tab:THAComp} and Fig.~\ref{fig:thaScatter}. The outcome distinctly displays a more accurate cup placement using the spatially-aware immersive AR system ($1.57^\circ \& 1.46^\circ$) compared to the SOP ($4.76^\circ~\&~4.77^\circ$), compared to the NI-AR system ($1.78^\circ~\&~1.43^\circ$) the abduction error is slightly less, whereas the anteversion error is marginally higher ($0.03^\circ$).  

For both procedures the deployment of our AR system lead to a comparable or higher accuracy, fewer X-ray images with a consequently lower radiation dose. For the total time, it has to be noted that our planning time does not include the recording of the X-ray images that were necessary to plan the procedure. This step however, as shown in~\cite{qian2017towards}, can be fully automated, resulting in an immediate availability of the fluoroscopic images. 

%%%%%%%%%%%%%%%%%%%%%%%%%%%%%%%%%%%%%%%%%%%%%%%%%%%%%%%%%%%%%
%%%%%%%%%%%%%%%%%%%%%%%%%%%%%%%%%%%%%%%%%%%%%%%%%%%%%%%%%%%%%

\section{Conclusion and Future Work}
We presented the embodiment of a novel interaction concept based on spatiotemporal-aware AR. For the two orthopedic use cases presented in this manuscript, our immersive AR system demonstrated improvements in time, number of X-ray acquisitions, radiation dose, and outcome during cup placement.

\begin{figure}
  \centering
  \includegraphics[width=\columnwidth]{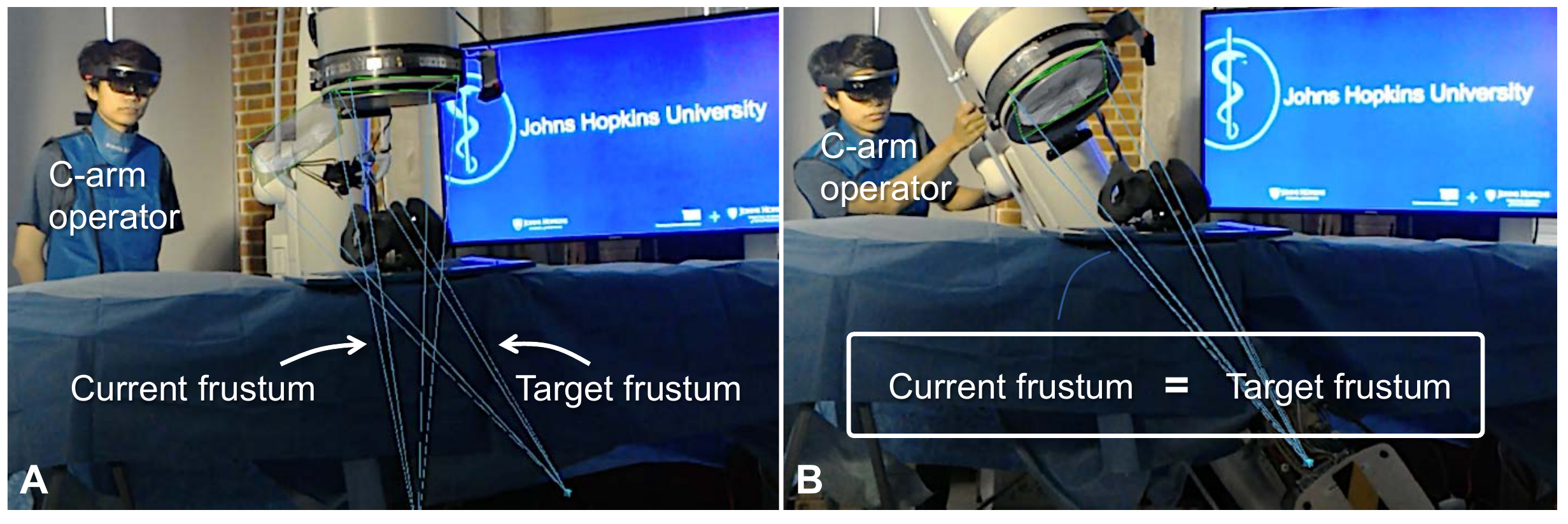}
  \caption{Visualization of a target frustum (\textbf{A}) allows the C-arm operator to align the current C-arm frustum with the surgeon's desired perspective (\textbf{B}) and eliminate the waste of time and radiation during fluoro hunting. This concept is an example of the capabilities of interactive frustums on moving information between different stake holders in the OR, \textit{i.e.} surgeon, patient, X-ray technician, staff, etc.}
    \label{fig:technicianInTheLoop}
\end{figure}

The spatiotemporal awareness inherent in AR overhauls the ill-posed communication between the surgeon, staff, and information; \textit{e.g.} Fig.~\ref{fig:technicianInTheLoop} shows the potential role of flying frustums and AR in effectively communicating desired X-ray views to the technician, eliminating unfavorable views and reducing the staff burnout.

An application of AR outside the OR is \textit{"surgical replay"}, where the  residents can review the surgery, accompanied with its temporal and spatial information including all the X-ray acquisitions and optical point-clouds from the patient site. This enables the medical trainees to identify distinct actions that were taken by the experienced surgeon based upon each image. Access to such 3D post-operative analysis has the potential to dramatically improve the quality of surgical education.

\begin{figure}
  \centering
  \includegraphics[width=\columnwidth]{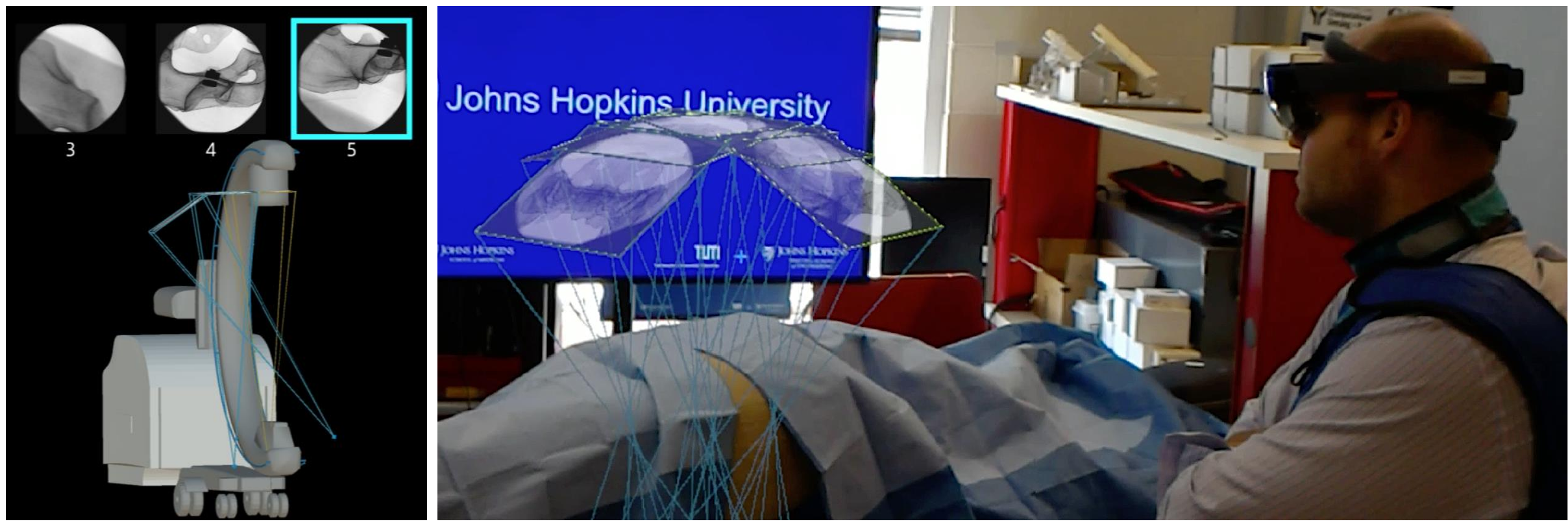}
  \caption{Spatial and temporal information from the surgery can be recorded and reviewed after surgery.}
    \label{fig:education}
\end{figure}

\begin{figure}
  \centering
  \includegraphics[width=\columnwidth]{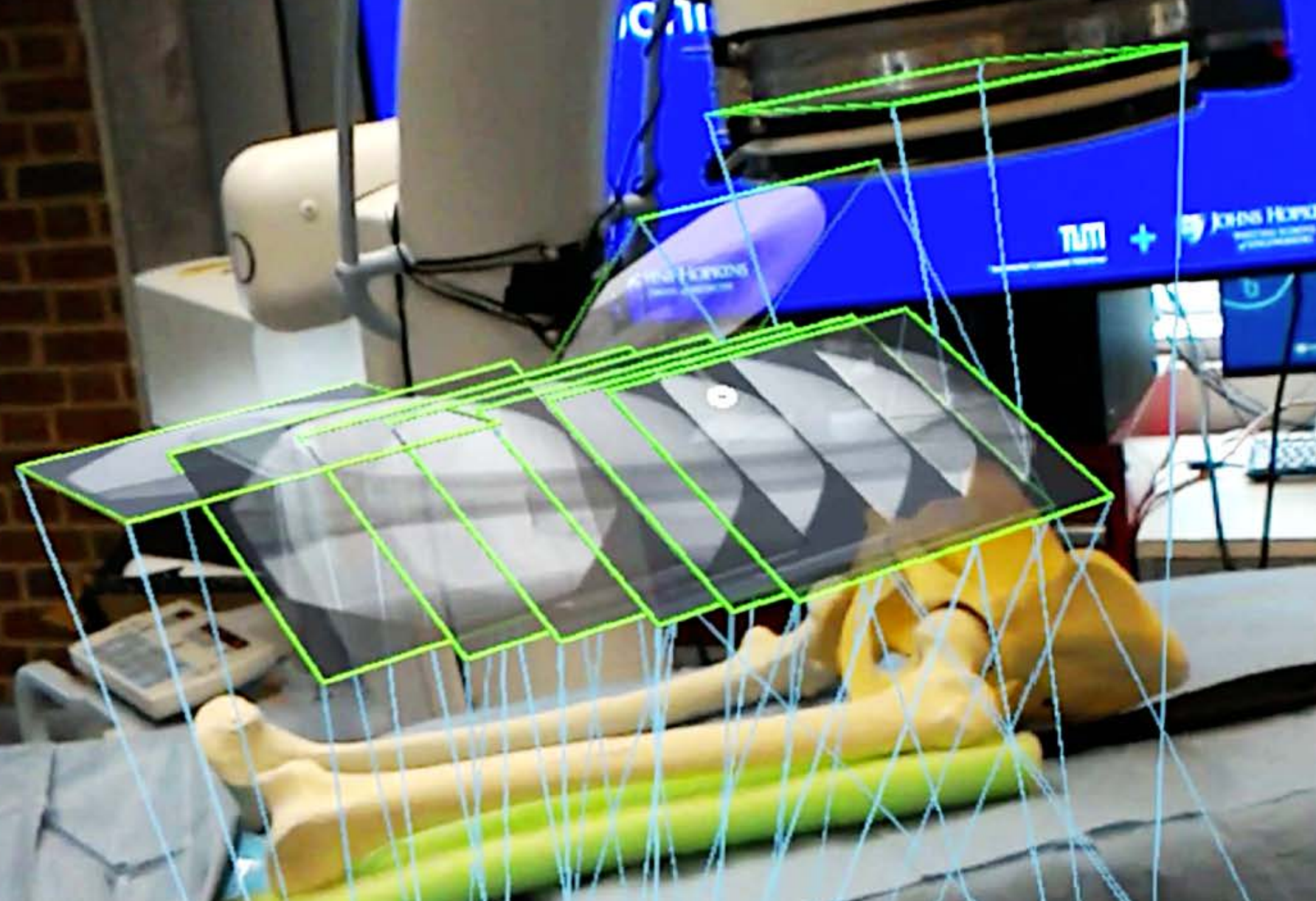}
  \caption{Interlocking of multiple X-ray frustums enables visualization of large anatomical structures.}
    \label{fig:legStitching}
\end{figure}

Direct visualization of X-ray images within their corresponding viewing frustums delivers intuition that effectively unites the content of the 2D image with the 3D imaged anatomy. In this setting, images from various perspectives can be grouped within their frustums to form multi- or extended-view representations of the anatomy. The interlocked frustums shown in Fig.~\ref{fig:legStitching} are examples for such visualization concept, that can particularly benefit interventions where leg-length discrepancies or malrotations in tibio- and lateral/distal-femoral angles are major concerns.

Though our solution delivers spatial awareness, it should not be regarded as a surgical navigation system. This is because marker-less tracking, currently, cannot deliver the level of accuracy achieved by marker-based surgical navigation or robotic systems. Our solution is merely an advanced visualization platform that enhances the interaction across the surgical ecosystem and promotes effective collaboration and team approach. 

The widespread adoption of human-centered AR in interventional routines requires careful considerations regarding surgeons' experience. The interaction with the virtual content using intuitive hand gestures, and resolving the perceptual ambiguities between real and virtual that occur due to vergence-accommodation conflict, can greatly contribute to the acceptance of AR technologies~\cite{fotouhi2019reflective}. Furthermore, development of artificial intelligence strategies can \textit{i)} create semantic understanding from the surgical environment and augment surgeon's intelligence, and \textit{ii)} enhance the spatial mapping and co-localization, thus improving the stability of marker-less AR systems. Lastly, future integration of eye tracking systems into HMDs can circumvent the internal eye-to-display calibration, adjust the rendering based on each user, and replace unnecessary gestures with gaze-based interaction.

As shown in this paper, the introduction of new technology requires a user-centric design for its full integration into the clinical workflow. The ultimate goal may be the discovery of new surgical workflows enabled by the introduction of novel technology. This goal could only be achieved with a close and extended partnership between surgeons and technical experts, as well as the full integration of the advanced technology in the broad spectrum of surgical teaching, training, planning, workflow, and documentation.

\bibliographystyle{ieeetran}

\begin{thebibliography}{10}

\bibitem{siewerdsen2005volume}
J.~Siewerdsen, D.~Moseley, S.~Burch, S.~Bisland, A.~Bogaards, B.~Wilson, and
  D.~Jaffray, ``Volume ct with a flat-panel detector on a mobile, isocentric
  c-arm: Pre-clinical investigation in guidance of minimally invasive
  surgery,'' \emph{Medical physics}, vol.~32, no.~1, pp. 241--254, 2005.

\bibitem{hott2004intraoperative}
J.~S. Hott, V.~R. Deshmukh, J.~D. Klopfenstein, V.~K. Sonntag, C.~A. Dickman,
  R.~F. Spetzler, and S.~M. Papadopoulos, ``Intraoperative iso-c c-arm
  navigation in craniospinal surgery: the first 60 cases,''
  \emph{Neurosurgery}, vol.~54, no.~5, pp. 1131--1137, 2004.

\bibitem{miller2010occupational}
D.~L. Miller, E.~Van{\'o}, G.~Bartal, S.~Balter, R.~Dixon, R.~Padovani,
  B.~Schueler, J.~F. Cardella, and T.~De~Ba{\`e}re, ``Occupational radiation
  protection in interventional radiology: a joint guideline of the
  cardiovascular and interventional radiology society of europe and the society
  of interventional radiology,'' \emph{Cardiovascular and interventional
  radiology}, vol.~33, no.~2, pp. 230--239, 2010.

\bibitem{theocharopoulos2003occupational}
N.~Theocharopoulos, K.~Perisinakis, J.~Damilakis, G.~Papadokostakis,
  A.~Hadjipavlou, and N.~Gourtsoyiannis, ``Occupational exposure from common
  fluoroscopic projections used in orthopaedic surgery,'' \emph{JBJS}, vol.~85,
  no.~9, pp. 1698--1703, 2003.

\bibitem{kim2008use}
C.~W. Kim, Y.-P. Lee, W.~Taylor, A.~Oygar, and W.~K. Kim, ``Use of
  navigation-assisted fluoroscopy to decrease radiation exposure during
  minimally invasive spine surgery,'' \emph{The Spine Journal}, vol.~8, no.~4,
  pp. 584--590, 2008.

\bibitem{kakarla2010placement}
U.~K. Kakarla, A.~S. Little, S.~W. Chang, V.~K. Sonntag, and N.~Theodore,
  ``Placement of percutaneous thoracic pedicle screws using neuronavigation,''
  \emph{World neurosurgery}, vol.~74, no.~6, pp. 606--610, 2010.

\bibitem{crawford2017surgical}
N.~R. Crawford, N.~Theodore, and M.~A. Foster, ``Surgical robot platform,''
  Oct.~10 2017, uS Patent 9,782,229.

\bibitem{yi2018robotic}
T.~Yi, V.~Ramchandran, J.~H. Siewerdsen, and A.~Uneri, ``Robotic drill guide
  positioning using known-component 3d--2d image registration,'' \emph{Journal
  of Medical Imaging}, vol.~5, no.~2, p. 021212, 2018.

\bibitem{joskowicz2016computer}
L.~Joskowicz and E.~J. Hazan, ``Computer aided orthopaedic surgery: incremental
  shift or paradigm change?'' 2016.

\bibitem{grupp2019pose}
R.~B. Grupp, R.~Hegeman, R.~Murphy, C.~Alexander, Y.~Otake, B.~McArthur,
  M.~Armand, and R.~H. Taylor, ``Pose estimation of periacetabular osteotomy
  fragments with intraoperative x-ray navigation,'' \emph{IEEE Transactions on
  Biomedical Engineering}, 2019.

\bibitem{goerres2017planning}
J.~Goerres, A.~Uneri, M.~Jacobson, B.~Ramsay, T.~De~Silva, M.~Ketcha, R.~Han,
  A.~Manbachi, S.~Vogt, G.~Kleinszig \emph{et~al.}, ``Planning, guidance, and
  quality assurance of pelvic screw placement using deformable image
  registration,'' \emph{Physics in Medicine \& Biology}, vol.~62, no.~23, p.
  9018, 2017.

\bibitem{synowitz2006surgeon}
M.~Synowitz and J.~Kiwit, ``Surgeon’s radiation exposure during percutaneous
  vertebroplasty,'' \emph{Journal of Neurosurgery: Spine}, vol.~4, no.~2, pp.
  106--109, 2006.

\bibitem{starr2001preliminary}
A.~Starr, A.~Jones, C.~Reinert, and D.~Borer, ``Preliminary results and
  complications following limited open reduction and percutaneous screw
  fixation of displaced fractures of the acetabulum.'' \emph{Injury}, vol.~32,
  pp. SA45--50, 2001.

\bibitem{aagaard2017interaction}
K.~Aagaard, B.~S. Laursen, B.~S. Rasmussen, and E.~E. S{\o}rensen,
  ``Interaction between nurse anesthetists and patients in a highly
  technological environment,'' \emph{Journal of PeriAnesthesia Nursing},
  vol.~32, no.~5, pp. 453--463, 2017.

\bibitem{laverdiere2019augmented}
C.~Laverdi{\`e}re, J.~Corban, J.~Khoury, S.~M. Ge, J.~Schupbach, E.~J. Harvey,
  R.~Reindl, and P.~A. Martineau, ``Augmented reality in orthopaedics: a
  systematic review and a window on future possibilities,'' \emph{The Bone \&
  Joint Journal}, vol. 101, no.~12, pp. 1479--1488, 2019.

\bibitem{hatscher2017gazetap}
B.~Hatscher, M.~Luz, L.~E. Nacke, N.~Elkmann, V.~M{\"u}ller, and C.~Hansen,
  ``Gazetap: towards hands-free interaction in the operating room,'' in
  \emph{Proceedings of the 19th ACM international conference on multimodal
  interaction}.\hskip 1em plus 0.5em minus 0.4em\relax ACM, 2017, pp. 243--251.

\bibitem{mewes2017touchless}
A.~Mewes, B.~Hensen, F.~Wacker, and C.~Hansen, ``Touchless interaction with
  software in interventional radiology and surgery: a systematic literature
  review,'' \emph{International journal of computer assisted radiology and
  surgery}, vol.~12, no.~2, pp. 291--305, 2017.

\bibitem{ma2016device}
M.~Ma, P.~Fallavollita, S.~Habert, S.~Weidert, and N.~Navab, ``Device-and
  system-independent personal touchless user interface for operating rooms: One
  personal ui to control all displays in an operating room.''
  \emph{International journal of computer assisted radiology and surgery},
  vol.~11, no.~6, pp. 853--861, 2016.

\bibitem{sato1998image}
Y.~Sato, M.~Nakamoto, Y.~Tamaki, T.~Sasama, I.~Sakita, Y.~Nakajima, M.~Monden,
  and S.~Tamura, ``Image guidance of breast cancer surgery using 3-d ultrasound
  images and augmented reality visualization,'' \emph{IEEE Transactions on
  Medical Imaging}, vol.~17, no.~5, pp. 681--693, 1998.

\bibitem{navab1999merging}
N.~Navab, A.~Bani-Kashemi, and M.~Mitschke, ``Merging visible and invisible:
  Two camera-augmented mobile c-arm (camc) applications,'' in \emph{Proceedings
  2nd IEEE and ACM International Workshop on Augmented Reality
  (IWAR'99)}.\hskip 1em plus 0.5em minus 0.4em\relax IEEE, 1999, pp. 134--141.

\bibitem{navab2009camera}
N.~Navab, S.-M. Heining, and J.~Traub, ``Camera augmented mobile c-arm (camc):
  calibration, accuracy study, and clinical applications,'' \emph{IEEE
  transactions on medical imaging}, vol.~29, no.~7, pp. 1412--1423, 2009.

\bibitem{fischer2016preclinical}
M.~Fischer, B.~Fuerst, S.~C. Lee, J.~Fotouhi, S.~Habert, S.~Weidert, E.~Euler,
  G.~Osgood, and N.~Navab, ``Preclinical usability study of multiple augmented
  reality concepts for k-wire placement,'' \emph{International journal of
  computer assisted radiology and surgery}, vol.~11, no.~6, pp. 1007--1014,
  2016.

\bibitem{fotouhi2016interventional}
J.~Fotouhi, B.~Fuerst, S.~C. Lee, M.~Keicher, M.~Fischer, S.~Weidert, E.~Euler,
  N.~Navab, and G.~Osgood, ``Interventional 3d augmented reality for orthopedic
  and trauma surgery,'' in \emph{16th Annual Meeting of the Int. Society for
  Computer Assisted Orthopedic Surgery (CAOS)}, 2016.

\bibitem{tucker2018towards}
E.~Tucker, J.~Fotouhi, M.~Unberath, S.~C. Lee, B.~Fuerst, A.~Johnson,
  M.~Armand, G.~M. Osgood, and N.~Navab, ``Towards clinical translation of
  augmented orthopedic surgery: from pre-op ct to intra-op x-ray via rgbd
  sensing,'' in \emph{Medical Imaging 2018: Imaging Informatics for Healthcare,
  Research, and Applications}, vol. 10579.\hskip 1em plus 0.5em minus
  0.4em\relax International Society for Optics and Photonics, 2018, p. 105790J.

\bibitem{fotouhi2018plan}
J.~Fotouhi, C.~P. Alexander, M.~Unberath, G.~Taylor, S.~C. Lee, B.~Fuerst,
  A.~Johnson, G.~M. Osgood, R.~H. Taylor, H.~Khanuja \emph{et~al.}, ``Plan in
  2-d, execute in 3-d: an augmented reality solution for cup placement in total
  hip arthroplasty,'' \emph{Journal of Medical Imaging}, vol.~5, no.~2, p.
  021205, 2018.

\bibitem{muller2019augmented}
F.~M{\"u}ller, S.~Roner, F.~Liebmann, J.~M. Spirig, P.~F{\"u}rnstahl, and
  M.~Farshad, ``Augmented reality navigation for spinal pedicle screw
  instrumentation using intraoperative 3d imaging,'' \emph{The Spine Journal},
  2019.

\bibitem{agten2018augmented}
C.~A. Agten, C.~Dennler, A.~B. Rosskopf, L.~Jaberg, C.~W. Pfirrmann, and
  M.~Farshad, ``Augmented reality--guided lumbar facet joint injections,''
  \emph{Investigative radiology}, vol.~53, no.~8, pp. 495--498, 2018.

\bibitem{gibby2019head}
J.~T. Gibby, S.~A. Swenson, S.~Cvetko, R.~Rao, and R.~Javan, ``Head-mounted
  display augmented reality to guide pedicle screw placement utilizing computed
  tomography,'' \emph{International journal of computer assisted radiology and
  surgery}, vol.~14, no.~3, pp. 525--535, 2019.

\bibitem{van2018augmented}
B.~van Duren, K.~Sugand, R.~Wescott, R.~Carrington, and A.~Hart, ``Augmented
  reality fluoroscopy simulation of the guide-wire insertion in dhs surgery: A
  proof of concept study,'' \emph{Medical engineering \& physics}, vol.~55, pp.
  52--59, 2018.

\bibitem{cho2018can}
H.~S. Cho, M.~S. Park, S.~Gupta, I.~Han, H.-S. Kim, H.~Choi, and J.~Hong, ``Can
  augmented reality be helpful in pelvic bone cancer surgery? an in vitro
  study,'' \emph{Clinical Orthopaedics and Related Research{\textregistered}},
  vol. 476, no.~9, pp. 1719--1725, 2018.

\bibitem{ogawa2018pilot}
H.~Ogawa, S.~Hasegawa, S.~Tsukada, and M.~Matsubara, ``A pilot study of
  augmented reality technology applied to the acetabular cup placement during
  total hip arthroplasty,'' \emph{The Journal of arthroplasty}, vol.~33, no.~6,
  pp. 1833--1837, 2018.

\bibitem{brun2018mixed}
H.~Brun, R.~Bugge, L.~Suther, S.~Birkeland, R.~Kumar, E.~Pelanis, and O.~Elle,
  ``Mixed reality holograms for heart surgery planning: first user experience
  in congenital heart disease,'' \emph{European Heart Journal-Cardiovascular
  Imaging}, 2018.

\bibitem{pelargos2017utilizing}
P.~E. Pelargos, D.~T. Nagasawa, C.~Lagman, S.~Tenn, J.~V. Demos, S.~J. Lee,
  T.~T. Bui, N.~E. Barnette, N.~S. Bhatt, N.~Ung \emph{et~al.}, ``Utilizing
  virtual and augmented reality for educational and clinical enhancements in
  neurosurgery,'' \emph{Journal of Clinical Neuroscience}, vol.~35, pp. 1--4,
  2017.

\bibitem{deib2018image}
G.~Deib, A.~Johnson, M.~Unberath, K.~Yu, S.~Andress, L.~Qian, G.~Osgood,
  N.~Navab, F.~Hui, and P.~Gailloud, ``Image guided percutaneous spine
  procedures using an optical see-through head mounted display: proof of
  concept and rationale,'' \emph{Journal of neurointerventional surgery},
  vol.~10, no.~12, pp. 1187--1191, 2018.

\bibitem{chimenti2015google}
P.~C. Chimenti and D.~J. Mitten, ``Google glass as an alternative to standard
  fluoroscopic visualization for percutaneous fixation of hand fractures: a
  pilot study,'' \emph{Plastic and reconstructive surgery}, vol. 136, no.~2,
  pp. 328--330, 2015.

\bibitem{moreta2018augmented}
R.~Moreta-Martinez, D.~Garc{\'\i}a-Mato, M.~Garc{\'\i}a-Sevilla,
  R.~P{\'e}rez-Ma{\~n}anes, J.~Calvo-Haro, and J.~Pascau, ``Augmented reality
  in computer-assisted interventions based on patient-specific 3d printed
  reference,'' \emph{Healthcare technology letters}, vol.~5, no.~5, pp.
  162--166, 2018.

\bibitem{meulstee2019toward}
J.~W. Meulstee, J.~Nijsink, R.~Schreurs, L.~M. Verhamme, T.~Xi, H.~H. Delye,
  W.~A. Borstlap, and T.~J. Maal, ``Toward holographic-guided surgery,''
  \emph{Surgical innovation}, vol.~26, no.~1, pp. 86--94, 2019.

\bibitem{ma2018three}
L.~Ma, Z.~Zhao, B.~Zhang, W.~Jiang, L.~Fu, X.~Zhang, and H.~Liao,
  ``Three-dimensional augmented reality surgical navigation with hybrid optical
  and electromagnetic tracking for distal intramedullary nail interlocking,''
  \emph{The International Journal of Medical Robotics and Computer Assisted
  Surgery}, vol.~14, no.~4, p. e1909, 2018.

\bibitem{andress2018fly}
S.~Andress, A.~Johnson, M.~Unberath, A.~F. Winkler, K.~Yu, J.~Fotouhi,
  S.~Weidert, G.~Osgood, and N.~Navab, ``On-the-fly augmented reality for
  orthopedic surgery using a multimodal fiducial,'' \emph{Journal of Medical
  Imaging}, vol.~5, no.~2, p. 021209, 2018.

\bibitem{hajek2018closing}
J.~Hajek, M.~Unberath, J.~Fotouhi, B.~Bier, S.~C. Lee, G.~Osgood, A.~Maier,
  M.~Armand, and N.~Navab, ``Closing the calibration loop: an
  inside-out-tracking paradigm for augmented reality in orthopedic surgery,''
  in \emph{International Conference on Medical Image Computing and
  Computer-Assisted Intervention}.\hskip 1em plus 0.5em minus 0.4em\relax
  Springer, 2018, pp. 299--306.

\bibitem{fotouhi2019co}
J.~Fotouhi, M.~Unberath, T.~Song, J.~Hajek, S.~C. Lee, B.~Bier, A.~Maier,
  G.~Osgood, M.~Armand, and N.~Navab, ``Co-localized augmented human and x-ray
  observers in collaborative surgical ecosystem,'' \emph{International journal
  of computer assisted radiology and surgery}, vol.~14, no.~9, pp. 1553--1563,
  2019.

\bibitem{fotouhi2019interactive}
J.~Fotouhi, M.~Unberath, T.~Song, W.~Gu, A.~Johnson, G.~Osgood, M.~Armand, and
  N.~Navab, ``Interactive flying frustums (iffs): spatially aware surgical data
  visualization,'' \emph{International journal of computer assisted radiology
  and surgery}, pp. 1--10, 2019.

\bibitem{hartley2003multiple}
R.~Hartley and A.~Zisserman, \emph{Multiple view geometry in computer
  vision}.\hskip 1em plus 0.5em minus 0.4em\relax Cambridge university press,
  2003.

\bibitem{alexander2020augmented}
C.~Alexander, A.~E. Loeb, J.~Fotouhi, N.~Navab, M.~Armand, and H.~S. Khanuja,
  ``Augmented reality for acetabular component placement in direct anterior
  total hip arthroplasty,'' \emph{The Journal of Arthroplasty}, 2020.

\bibitem{lewinnek1978dislocations}
G.~E. Lewinnek, J.~Lewis, R.~Tarr, C.~Compere, and J.~Zimmerman, ``Dislocations
  after total hip-replacement arthroplasties.'' \emph{The Journal of bone and
  joint surgery. American volume}, vol.~60, no.~2, pp. 217--220, 1978.

\bibitem{qian2017towards}
L.~Qian, M.~Unberath, K.~Yu, B.~Fuerst, A.~Johnson, N.~Navab, and G.~Osgood,
  ``Towards virtual monitors for image guided interventions-real-time streaming
  to optical see-through head-mounted displays,'' \emph{arXiv preprint
  arXiv:1710.00808}, 2017.

\bibitem{fotouhi2019reflective}
J.~Fotouhi, T.~Song, A.~Mehrfard, G.~Taylor, A.~Martin-Gomez, B.~Fuerst,
  M.~Armand, M.~Unberath, and N.~Navab, ``Reflective-ar display: An interaction
  methodology for virtual-real alignment in medical robotics,'' \emph{arXiv
  preprint arXiv:1907.10138}, 2019.

\end{thebibliography}

\end{document}